\documentclass{article}

\usepackage[preprint]{neurips_2026}

\usepackage[utf8]{inputenc} 
\usepackage[T1]{fontenc}    
\usepackage{hyperref}       
\usepackage{url}            
\usepackage{booktabs}       
\usepackage{amsfonts}       
\usepackage{nicefrac}       
\usepackage{microtype}      
\usepackage{xcolor}         

\usepackage{tabularx}
\usepackage{amsmath}
\usepackage{makecell}
\usepackage{array} 
\usepackage{multirow}
\usepackage{lipsum}
\usepackage{booktabs}
\usepackage{color}
\usepackage{colortbl}
\usepackage{subcaption}

\newcommand{\gray}[1]{{\color{gray}#1}}

\newcommand{\paragrapht}[1]{\noindent\textbf{#1}}  

\usepackage{pifont}
\usepackage{amssymb}

\usepackage{xspace}
\usepackage{array}
\usepackage{layouts}
\usepackage{algorithm}
\usepackage{bbding}
\usepackage{listings}
\usepackage{pifont}
\usepackage{wasysym}
\usepackage{float}
\usepackage{soul}
\usepackage{footnote}
\makesavenoteenv{tabular}
\makesavenoteenv{table}
\makesavenoteenv{figure}
\usepackage{pythonhighlight}
\usepackage{enumitem}
\usepackage{graphicx}
\usepackage{wrapfig}

\usepackage{hyperref}
\usepackage[capitalize,noabbrev]{cleveref}
\usepackage{tikz}
\usepackage{pgfplots}
\usepackage{threeparttable}
\usetikzlibrary{patterns,patterns.meta}

\newcommand{\method}{SPARK-VLA\xspace}%
\newcommand{\methodlong}{\textbf{Spa}tial \textbf{R}epresentation distillation and \textbf{K}nowledge routing for \textbf{VLA}s\xspace}%

\newcommand{\ie}{\textit{i.e.}}
\newcommand{\eg}{\textit{e.g.}}

\newcommand{\tabref}[1]{Table~\ref{#1}}
\newcommand{\figref}[1]{Figure~\ref{#1}}

\definecolor{citeblue}{RGB}{35, 90, 160}
\definecolor{urlblue}{RGB}{25, 85, 155}

\hypersetup{
    colorlinks=true,
    citecolor=blue,
    linkcolor=blue,
    urlcolor=blue
}

\title{Spatial Representation Distillation and Knowledge Routing for Vision-Language-Action Models}

\author{%
  Jungin Park\textsuperscript{1,2,*}
  \And
  Chaoran Zhu\textsuperscript{3,*}
  \And
  Changjae Oh\textsuperscript{3}
  \And
  Kwanghoon Sohn\textsuperscript{2,$\dagger$}
}

\begin{document}

\maketitle

\makeatletter
\begingroup
\renewcommand{\thefootnote}{}
\renewcommand{\@makefntext}[1]{\noindent #1}
\footnotetext{%
  $^{1}$ DGIST, South Korea. \quad
  $^{2}$ Yonsei University, South Korea. \quad
  $^{3}$ Queen Mary University of London, UK.

  \par\noindent
  $^{*}$ Equally contributed. \quad
  $^{\dagger}$ Corresponding author: \texttt{khsohn@yonsei.ac.kr}
}
\endgroup
\makeatother

\begin{abstract}
    Understanding three-dimensional physical structure has become essential for vision-language-action models (VLAs). 
    However, existing approaches either rely on explicit geometric inputs at inference time or adopt a single visual pathway, potentially compromising the visual-semantic knowledge inherited from pre-trained models.
    We present \method, \methodlong, a preserve-specialize-route framework for spatially adapting VLAs without requiring external 3D inputs or experts at inference.
    \method consists of two key components: 
    (1) spatial representation distillation, which transfers 3D understanding from a frozen 3D expert to a LoRA-adapted visual branch by aligning attention responses while preserving the original visual pathway, and (2) a lightweight knowledge router, which selectively forwards action-relevant visual tokens to the language model using routing supervision derived from action-loss gradients.
    Across LIBERO, LIBERO-Plus, CALVIN, and real-world manipulation, \method achieves strong performance with only a 0.5B language backbone and approximately 53M trainable parameters, providing an effective and parameter-efficient approach to 3D-aware VLA learning.
    Code is available at \url{https://github.com/park-jungin/spark-vla}.
    
\end{abstract}
\section{Introduction}
    Vision-language-action models (VLAs) learn to generate robot actions from visual observations and language instructions, offering a unified approach to perception, instruction understanding, and control.
    Recent systems have benefited from large-scale vision-language pre-training, robot demonstrations, and increasingly expressive action decoders~\citep{rt-2, openvla, pi0, dexvla, cot-vla}.
    Nevertheless, recognizing the objects involved in an instruction does not necessarily provide the geometric information required to manipulate them.
    Reaching, grasping, and placing an object also depend on its position, orientation, extent, and spatial relationship to the robot and surrounding objects.
    Even with multiple camera views, aggregating 2D visual features insufficiently yields geometrically consistent representations across viewpoints~\citep{mvsnet,epipolar-transformers,rvt}. 
    In other words, understanding and acting in the physical world is a fundamental capability for embodied intelligence, requiring joint modeling of perception, language, action, and \emph{three-dimensional spatial structure}.

    Recent studies have increasingly suggested that spatial awareness remains a key bottleneck for reliable robot manipulation and have investigated explicit 3D geometric priors, such as depth maps~\citep{3d-cavla}, point clouds~\citep{geovla}, or other geometric cues~\citep{spatialvla, tracevla}, and implicit alignment between 2D and 3D representations~\citep{evo-0, spatial-forcing} with pre-trained vision-language models (VLMs).
    Despite their promise, both directions have important limitations:
    (1) Explicit 3D integration typically depends on extra sensing or reconstructed geometric inputs that are not uniformly available across robot platforms and datasets, and (2) implicit alignment often distorts the original 2D representation space, which can weaken the perceptual capability inherited from strong vision-language backbones~\citep{chatvla-2, evo-1}.

    \begin{figure}[t]
        \centering
        \begin{subfigure}[b]{0.49\linewidth}
            \includegraphics[width=\linewidth]{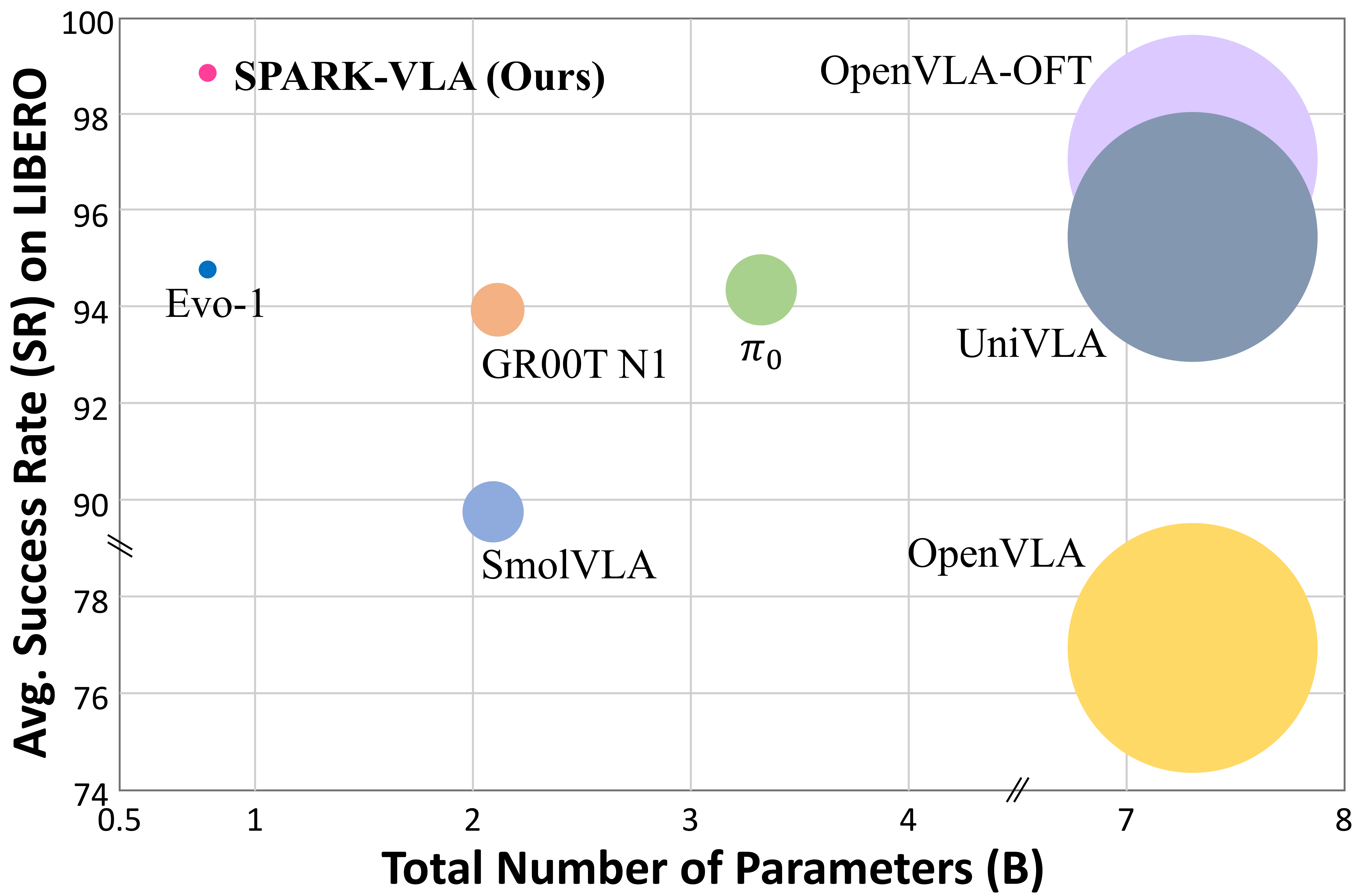}
            \caption{Performance comparison on LIBERO}
            \label{fig:libero}
        \end{subfigure}
        \hfill
        \begin{subfigure}[b]{0.49\linewidth}
            \includegraphics[width=\linewidth]{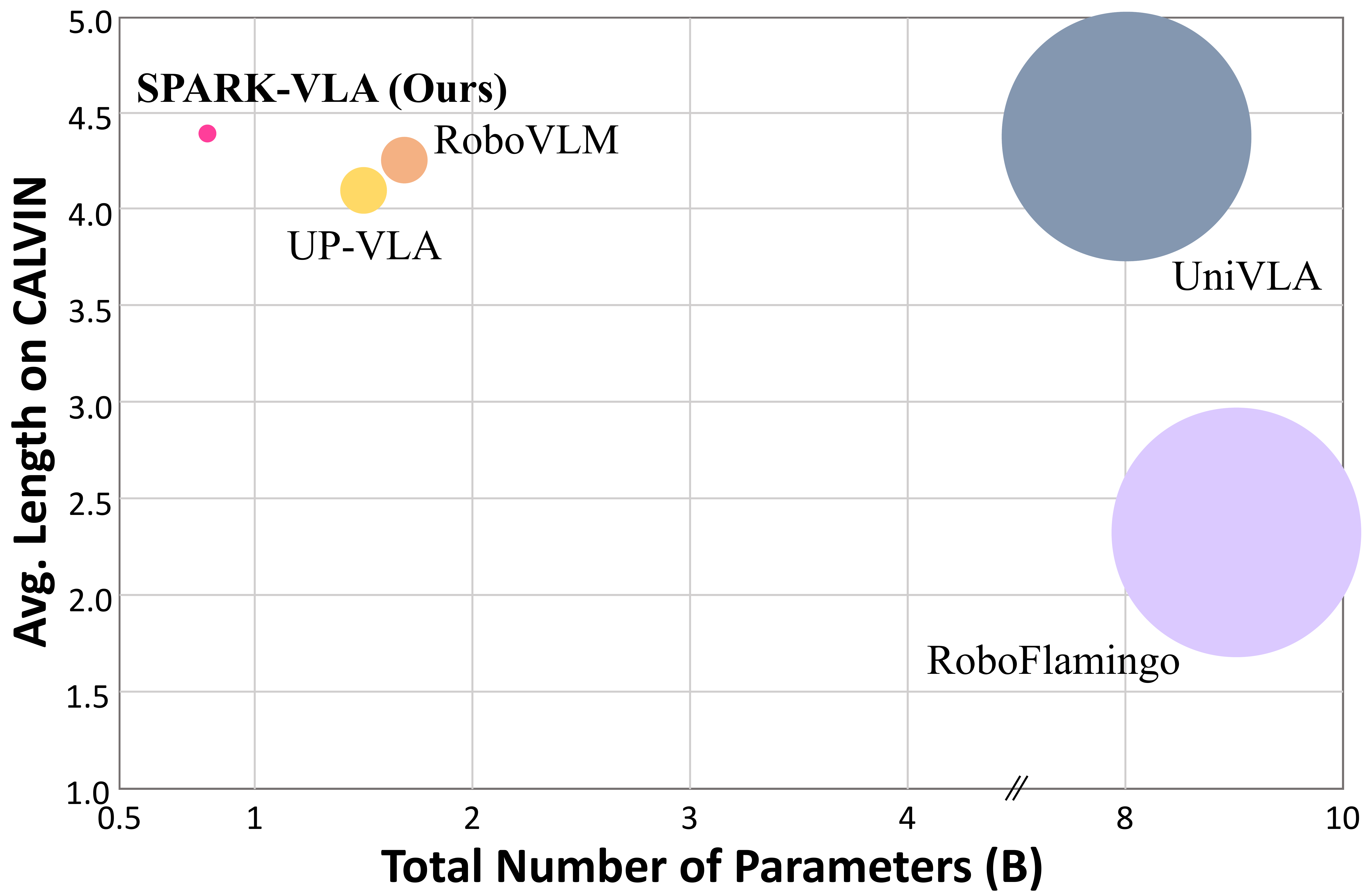}
            \vspace{-1.0em}
            \caption{Performance comparison on CALVIN (ABC$\rightarrow$D)}
            \label{fig:calvin}
        \end{subfigure}
        \caption{
        We depict (a) the averaged success rate for LIBERO and (b) the average length for CALVIN, with respect to the total number of parameters. The center point and the size of circles indicates the averaged performance and the size of LLMs in each model, respectively.}
        \label{fig:performance}
    \end{figure}

    
    We propose \method (\methodlong), a framework that enhances the 3D spatial understanding of a pre-trained VLM while preserving its original perceptual capability. 
    Specifically, \method establishes a spatial pathway with low-rank adapters~\citep{lora} \emph{parallel} to the original visual pathway within the pre-trained vision backbone.
    During training, the original pathway remains frozen, while the adapted pathway learns geometry-aware representations from a pre-trained 3D experts (\eg, VGGT~\citep{vggt}) through attention response distillation~\citep{v-lynx}: the adapted spatial query is shared with the expert's keys and values, allowing the spatial pathway to learn the geometry-aware information aggregated for the query.

    \method further introduces an \emph{action-driven knowledge router} to combine these complementary pathways.
    The router predicts instruction-conditioned scores over generic and spatial tokens and selectively forwards informative tokens to the language model, determining whether each region is represented by generic tokens, spatial tokens, both, or neither.
    To train the router without manually annotated supervision, we derive routing targets from the gradients of the action loss with respect to the router gates that indicate the usefulness of the tokens for action prediction.
    Together, preservation of the original pathway, specialization through spatial distillation, and action-driven routing provide a unified approach to integrating geometric knowledge into pre-trained VLAs.    

    Experimental results on simulation benchmarks, including LIBERO~\citep{libero} and CALVIN~\citep{calvin}, show that \method with only a 0.5B language backbone and 53M trainable parameters achieves performance comparable to or even better than substantially larger VLAs, as shown in \figref{fig:performance}.
    In addition, \method generalizes robustly to challenging settings such as LIBERO-Plus~\citep{libero-plus} without any additional training.
    Experiments in real-world environments further demonstrate that our model achieves high sample efficiency under data-constrained conditions, outperforming much larger baselines.
\section{Related Work}
\paragrapht{Vision-language-action models (VLAs).}
    VLAs have emerged as a promising paradigm for generalist robot learning by jointly modeling visual perception, language understanding, and action generation~\citep{palme, rt-2, openvla, octo, robo-flamingo, openx, pi0}.
    Recent advances have explored diverse action generation strategies, including autoregressive action tokenization~\citep{openvla,pi0-fast}, diffusion-based policies~\citep{diffusion-policy,dexvla,dita}, and flow matching~\citep{pi0}.
    Alongside these developments, several studies have investigated efficient adaptation and compact architectures~\citep{openvla-oft,vla-adapter,smolvla,nora}, as well as reasoning and long-horizon manipulation~\citep{cot-vla,chatvla-2,univla-iclr}.
    Despite their progress, transferring visual-semantic knowledge to robotic control does not necessarily provide the fine-grained geometric information required for precise manipulation.
    We addresses this gap through implicit spatial knowledge adaptation.

\paragrapht{Spatial understanding in VLAs.}
    Spatial reasoning is essential for robotic manipulation, particularly when actions depend on object geometry, relative positions, and physical interactions.
    Existing approaches have incorporated explicit geometric information, such as depth and 3D representations, to improve spatial understanding~\citep{spatialvla, tracevla, 3d-cavla, geovla}.
    More recently, several studies have explored implicit spatial learning by transferring knowledge from pre-trained geometric models or aligning visual representations with spatial priors~\citep{spatial-forcing, evo-0, geoaware-vla}.
    Other approaches have emphasized preserving pre-trained semantic representations during efficient VLA adaptation~\citep{chatvla-2, evo-1}.
    However, incorporating geometric knowledge into an existing visual pathway introduces potentially competing requirements: acquiring spatial information while retaining generic visual-semantic knowledge.
    Our \method preserves the original visual pathway and specializes a parallel LoRA-adapted pathway through attention response distillation~\citep{v-lynx}, allowing spatial specialization without modifying the preserved representation.

\paragrapht{Visual token selection and routing for action generation.}
    Recent work has shown that selective visual processing is important for efficient and effective action generation in VLAs~\citep{vla-cache, sp-vla, cog-vla, semantic-vla, spec-prune-vla, vla-pruner}.
    Existing methods exploit temporal redundancy through token caching and pruning~\citep{vla-cache,vla-pruner}, identify task-relevant information through instruction-conditioned selection~\citep{cog-vla,semantic-vla}, or incorporate action-aware importance into token reduction~\citep{spec-prune-vla,sp-vla}.
    These approaches primarily focus on identifying informative tokens within a given visual representation.
    In contrast, SPARK-VLA addresses a different selection problem: determining which knowledge source should represent each spatial position when both generic and spatially specialized features are available.
    Our action-driven router jointly ranks tokens from the two pathways using supervision derived from downstream action loss.
    
\section{Method}

\subsection{Motivation and overview}
    
    \begin{wrapfigure}{r}{0.55\textwidth}
    \centering
    \vspace{-10pt}
    \includegraphics[width=0.55\textwidth]{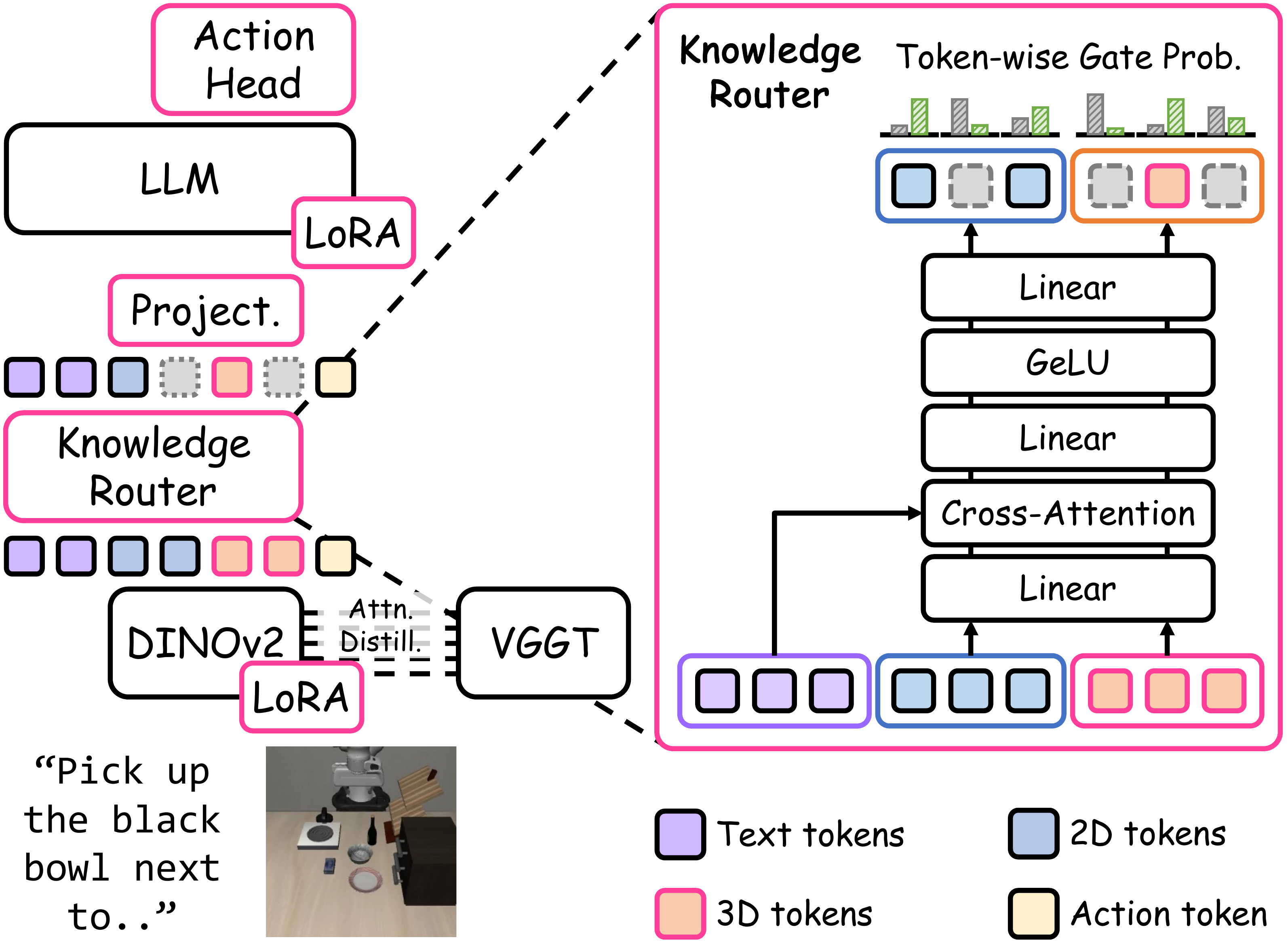}
    \caption{Overview of \method. Trainable modules are represented in the red boxes.}
    \label{fig:overall}
    \vspace{-10pt}
    \end{wrapfigure}

    Existing efforts to enhance vision-language-action models (VLAs) with spatial intelligence generally follow two directions: introducing explicit 3D inputs, such as depth maps or point clouds, or implicitly injecting spatial knowledge into the visual representation. 
    While the former often depends on additional sensing or reconstructed 3D signals, the latter can weaken the generic visual-semantic capability inherited from strong pre-trained vision-language backbones by forcing a single visual pathway to absorb both general perception and spatial reasoning. 
    
    To address this issue, we propose \method, which preserves the original visual pathway while introducing a lightweight spatially adapted pathway. 
    During training, we use a pre-trained 3D expert (\eg, VGGT~\citep{vggt}) to distill spatial knowledge into the adapted branch. 
    To effectively use the two branches, we further introduce a lightweight knowledge router that takes visual tokens, conditions on the language instruction, and predicts token-wise gate probabilities indicating which tokens are useful for action generation. 
    The selected tokens are then forwarded to the language model for action prediction, allowing \method to exploit complementary generic and spatial cues while suppressing redundant or noisy representations. 
    Note that the 3D expert is used only during training and is discarded at inference time.
    The overall architecture of our \method is shown in \figref{fig:overall}.

\subsection{Spatial representation distillation}
    
    To incorporate spatial understanding into the VLA's visual encoder without disrupting its original visual-semantic capability, we introduce a spatially adapted branch built on the pre-trained visual encoder.

    \paragrapht{Dual-branch visual encoding.}
    Let $f_\theta^v$ denote the visual encoder of a pre-trained vision-language model (\eg, Qwen2.5-VL~\citep{qwen25}) and $f_{\theta+\Delta\theta}^v$ denote its LoRA-adapted counterpart, where only the low-rank parameters $\Delta\theta$ are trainable.
    Given an RGB observation $\mathbf{I}_t$, \method extracts two complementary representations:
    \begin{equation}
        \mathbf{Z}_v = f_\theta^v(\mathbf{I}_t), \qquad
        \mathbf{Z}_p = f_{\theta+\Delta\theta}^v(\mathbf{I}_t),
    \end{equation}
    where $\mathbf{Z}_v$ preserves generic visual-semantic knowledge and $\mathbf{Z}_p$ is adapted to capture spatial reasoning.    
    During training, we further employ a pre-trained 3D encoder $f_\phi^p$ as a frozen teacher that provides spatial supervision to the adapted branch.
    In this way, the original branch preserves the pre-trained visual representation, while the adapted branch learns to encode additional spatial knowledge.

    \paragrapht{Attention response distillation.}
    Instead of directly forcing the adapted branch to mimic the output representation of the 3D teacher, we distill \emph{attention responses}~\citep{v-lynx}.
    Our key idea is to preserve compatibility with the original transformer structure by teaching the adapted branch \emph{how to attend} in a spatially informed manner.
    At the $l$-th layer of $f_{\theta+\Delta\theta}^v$, we obtain the query, key, and value embeddings, denoted by $Q_v^{(l)}$, $K_v^{(l)}$, and $V_v^{(l)}$.
    From the frozen 3D teacher, we extract the corresponding key and value embeddings, denoted by $K_p^{(l)}$ and $V_p^{(l)}$.
    We then compute two attention responses using the same query embedding $Q_v^{(l)}$:
    \begin{equation}
        \mathbf{A}_v^{(l)} = \mathrm{Attn}(Q_v^{(l)}, K_v^{(l)}, V_v^{(l)}),
        \qquad
        \mathbf{A}_p^{(l)} = \mathrm{Attn}(Q_v^{(l)}, K_p^{(l)}, V_p^{(l)}),
    \end{equation}
    where $\mathrm{Attn}(\cdot)$ denotes the scaled dot-product attention.
    Using the same query while replacing the key-value pairs allows the teacher response to specify how the adapted visual branch should interact with the attention mechanism to remain compatible with spatial knowledge.
    
    The attention response distillation objective for spatial representations is defined as:
    \begin{equation}
        \mathcal{L}_{\text{attn}}
        =
        \sum_l
        \left\|
        \mathbf{A}_v^{(l)} - \mathbf{A}_p^{(l)}
        \right\|_1,
    \end{equation}
    where $\|\cdot\|_1$ denotes the $\ell_1$ loss.
    By aligning attention responses across layers, the adapted branch acquires geometry-aware behavior from the 3D teacher while retaining the original visual pathway for generic scene understanding.

\subsection{Knowledge routing}
    While $\mathbf{Z}_v$ and $\mathbf{Z}_p$ provide complementary generic and spatial cues, not all tokens are equally useful for a given instruction and action prediction. 
    Directly feeding all tokens from both branches into the LLM may introduce redundant or noisy information, especially when spatially adapted tokens are irrelevant to the current instruction. 
    We therefore introduce a lightweight knowledge router that feeds instruction- and action-relevant visual tokens into the LLM.

\paragrapht{Router architecture.}
    We first collect the visual tokens from the generic and spatial branches and denote the merged token set by
    \begin{equation}
        \mathbf{Z} = \left[\mathbf{Z}_v ; \mathbf{Z}_p\right]
        = \{\mathbf{z}_n\}_{n=1}^{N},
    \end{equation}
    where $N$ is the total number of visual tokens.
    As shown in \figref{fig:overall}, the knowledge router consists of (1) a linear layer to project $\mathbf{Z}$ into the same dimensional space with the text embeddings $\mathbf{H}=\{\mathbf{h}_m\}_{m=1}^{M}$; (2) a cross-attention layer to compute the correlations between the visual and text representations; and (3) a lightweight MLP (Linear-GeLU-Linear).
    The output of the last linear layer followed by a Sigmoid activation serves as the gate probabilities $\mathbf{p}$:
    \begin{equation}
        \mathbf{p} = \sigma(\mathrm{Router}(\mathbf{Z}, \mathbf{H})/\tau) = \{p_n\}_{n=1}^N,
    \end{equation}
    where $\sigma(\cdot)$ is a sigmoid function and $\tau$ is a temperature parameter to control the sensitivity.
    
\paragrapht{Token selection and knowledge fusion.}
    Once the gate probabilities $\mathbf{p}$ are obtained, we select the top-$k$ tokens among all $N$ tokens:
    \begin{equation}
        \mathcal{S} = \mathrm{TopK}(\mathbf{p}, k), \qquad
        k=\max(k_{\min}, \mathrm{round}(\rho N)),
    \end{equation}
    where $\mathrm{TopK}(\mathbf{p}, k)$ returns the indices of the $k$ largest gate probabilities, $\rho$ indicates the routing ratio, and $k_{\min}$ prevents overly sparse routing.
    We then project the selected tokens to the LLM's input space by fusing them according to their spatial positions.
    At each spatial position, the routed representation is constructed from the selected tokens among its generic and spatial pair. 
    When both tokens are selected, their embeddings are concatenated and projected into the LLM input space. 
    If only one token is selected, it is concatenated with a zero embedding of the same dimension and passed through the same projection. 
    Positions for which neither token is selected are discarded. 
    This position-aware fusion preserves the alignment between generic and spatial knowledge, while producing a compact set of visual tokens with a unified input format for the LLM.
    

\paragrapht{Gradient-based gate supervision.}
    The main challenge in training the knowledge router is that there is no supervision signal indicating which visual tokens should be preserved.
    Instead of relying on heuristic salience measures~\citep{spec-prune-vla, vla-pruner} or learning token selection only through downstream action optimization~\citep{cog-vla, semantic-vla}, we derive routing targets from the gradient of the action loss with respect to the router gate probabilities.
    Given the action loss $\mathcal{L}_\text{act}$, the gradient of $\mathcal{L}_\text{act}$ \textit{w.r.t.} the gate probability $p_n$, \ie, $g_n = \partial \mathcal{L}_{\text{act}}/{\partial p_n}$,
    measures the influences of the $n$-th gate on action prediction.
    A negative gradient indicates that increasing the gate probability would reduce the action loss, suggesting that the corresponding token is useful.
    Therefore, we define the importance score as
    \begin{equation}
        s_n = \mathrm{ReLU}(-g_n).
    \end{equation}
    Since only routed tokens are forwarded to the LLM, the instantaneous gradients provide supervision only for the currently selected subset. 
    We therefore maintain a momentum-based moving average of token importance scores, allowing tokens that are not selected in the current step to still contribute to router training through their accumulated utility. 
    The detailed update rule is provided in \cref{app:hyperparameter}.

    From $\bar{s}_n$, we construct a gating target $\mathbf{y}=\{y_n\}_{n=1}^{N}$:
    \begin{equation}
        y_n =
        \begin{cases}
            1, & n \in \mathrm{TopK}(\bar{\mathbf{s}}, k),\\
            0, & \text{otherwise},
        \end{cases}
    \end{equation}
    where $\bar{\mathbf{s}}=\{\bar{s}_n\}_{n=1}^{N}$ denotes the momentum-updated importance score.
    This gating target enables the router to learn action-aware token selection without explicit token-level annotations.


\paragrapht{Routing objectives.}
    We train the knowledge router with the classification problem.
    To handle the imbalance between selected and discarded tokens, we adopt the focal binary cross-entropy loss~\citep{focal}:
    \begin{equation}
        \mathcal{L}_{\text{cls}}
        =\frac{1}{N}\sum_{n}\alpha_{y_{n}}(1-p_t)^\gamma
        \Big[-y_{n}\log p_{n}-(1-y_{n})\log(1-p_{n})\Big],
    \end{equation}
    where $p_t = y_{n}p_{n} + (1-y_{n})(1-p_{n})$.
    To further stabilize routing behavior, we additionally impose a budget regularization term that encourages the average routing ratio to match the target ratio $\rho$ and an entropy regularization term that controls the uncertainty of the gate probabilities:
    \begin{equation}
        \mathcal{L}_{\text{bud}}
        =\left|\frac{1}{N}\sum_n p_{n}-\rho\right|,
        \qquad
        \mathcal{L}_{\text{ent}}
        =-\frac{1}{N}\sum_{n}\Big[p_{n}\log p_{n}+(1-p_{n})\log(1-p_{n})\Big].
    \end{equation}

    The final router objective is defined as
    \begin{equation}
        \mathcal{L}_{\text{router}}
        =\mathcal{L}_{\text{cls}}
        +\lambda_{\text{bud}}\mathcal{L}_{\text{bud}}
        +\lambda_{\text{ent}}\mathcal{L}_{\text{ent}},
    \end{equation}
    where $\lambda_{\text{bud}}$ and $\lambda_{\text{ent}}$ control the impact of budget constraint and entropy regularization, respectively.
    By combining the knowledge routing with gradient-based pseudo supervision, the proposed knowledge router learns to preserve action-relevant visual tokens from both the generic and spatial branches while suppressing redundant or noisy representations.

\begin{table*}[t]
    \centering
    \small
    \caption{Comparisons with state-of-the-art methods on LIBERO. * incorporate extra depth or point cloud information from other sensors. Bold indicates the best performances among tiny-scale models.
    }\label{tab:libero}
    \setlength{\tabcolsep}{4pt}
        \begin{tabular}{lccccccc}
        \toprule
         \multirow{2}[2]{*}{Method}   &  Spatial    &   Object  &   Goal    &   Long    &   Average & \multirow{2}[2]{*}{\makecell{LLM \\ Size}}    &    \multirow{2}[2]{*}{\makecell{Train. \\ Params.}}   \\
         \cmidrule(lr){2-2} \cmidrule(lr){3-3} \cmidrule(lr){4-4} \cmidrule(lr){5-5} \cmidrule(lr){6-6}
                &   SR (\%)      &   SR (\%)     &   SR (\%)     &   SR (\%)     &   SR (\%)  &  \\
        \midrule
        \multicolumn{3}{l}{\hspace{-6pt}\gray{\textit{Large-scale models $\geq$ 7B}}} \\ 
        Octo~\citep{octo} &   78.9    &   85.7    &   84.6    &   51.1    &   75.1  & - & 93M   \\
        OpenVLA~\citep{openvla}     &   84.9    &   88.4    &   79.2    &   53.7    &   76.5    &   7B  &  111M  \\
        OpenVLA-OFT~\citep{openvla-oft} &   97.6    &   98.4    &   97.9    &   94.5    &   97.1    &   7B  & 111M \\
        Spatial Forcing~\citep{spatial-forcing}     &   99.4    &    99.6    &   98.8    &   96.0    &   98.5    &   7B  &  291M   \\
        UniVLA~\citep{univla-iclr}     &   95.4    &   99.0    &   93.6    &   94.0    &   95.5    &   7B  &  8B   \\
        CogVLA~\citep{cog-vla}     &   98.6    &   98.8    &   96.6    &   95.4    &   97.4    &  7B   &  98M   \\
        \midrule
        \multicolumn{3}{l}{\hspace{-6pt}\gray{\textit{1B $\leq$ Small-scale models $<$ 7B}}} \\ 
        4D-VLA~\citep{4d-vla}   &   88.9    &   95.2    &   90.9    &   79.1    &   88.6    &   3.8B  &   4.2B  \\
        $\pi_0$~\citep{pi0}     &   96.8    &   98.8    &   95.8    &   85.2    &   94.2  & 2B &  3.3B \\
        $\pi_0$-FAST~\citep{pi0-fast}    &   96.4    &   96.8    &   88.6    &   60.2    &   85.5  &  2B  & 3.3B \\
        SpatialVLA~\citep{spatialvla}  &   88.2    &   89.9    &   78.6    &   55.5    &   78.1  & 3B &  3.4B \\
        NORA~\citep{nora}   &   92.2    &   95.4    &   89.4    &   74.6    &   87.9    & 3B  & 4B  \\
        SmolVLA~\citep{smolvla}    &   93.0    &   94.0    &   91.0    &   77.0    &   88.8    & 1.7B  &  450M \\
        GR00T N1~\citep{groot}   &   94.4    &   97.6    &   93.0    &   90.6    &   93.9    & 1.5B  & 700M  \\
        VLANeXt~\citep{vlanext} &   99.0   &   99.2   &   96.6   &   94.6   &   97.4   &   2B  &  2.7B \\ 
        \midrule
        \multicolumn{3}{l}{\hspace{-6pt}\gray{\textit{Tiny-scale models $<$ 1B}}} \\ 
        Diffusion Policy~\citep{diffusion-policy}    &   78.3    &   92.5    &   68.3    &   50.5    &   72.4  & - &  263M  \\
        VLA-Adapter~\citep{vla-adapter}     &   99.6    &   \bf 99.6    &   \bf 98.2    &   96.4    &   \bf 98.5    &   0.5B  & 598M    \\
        Evo-1~\citep{evo-1}       &   92.7    &   97.7    &  96.3    &   92.3    &   94.8    &   0.5B    &   770M    \\
        \method (\textbf{Ours}) & \bf 99.6  &   \bf 99.6   & 98.0    &  \bf 96.6   & \bf 98.5  &   0.5B    &   53M    \\
        \bottomrule
        \end{tabular}


    
    \end{table*}

\subsection{Training objective}
    \method is trained with three objectives: the action loss, the spatial representation distillation loss, and the knowledge router loss:
    \begin{equation}\label{eq:objective}
        \mathcal{L}_{\text{spark}}
        =
        \mathcal{L}_{\text{act}}
        + \lambda_{\text{attn}} \mathcal{L}_{\text{attn}}
        + \lambda_{\text{router}} \mathcal{L}_{\text{router}},
    \end{equation}
    where $\lambda_{\text{attn}}$ balances the contribution of the spatial distillation term.
    We use $\ell_1$ regression for the action loss.
    Please see \cref{app:training-detail} for training details.

\section{Simulation Experiments}

\subsection{Experimental setting}
    \paragrapht{Dataset.}
    We evaluate \method on three benchmarks, including LIBERO~\citep{libero}, LIBERO-Plus~\citep{libero-plus}, and CALVIN~\citep{calvin}.
    \begin{itemize}[leftmargin=*]
    \item \textbf{LIBERO}~\citep{libero} is a widely used benchmark for lifelong robot manipulation that contains four task suites, namely LIBERO-Spatial, LIBERO-Object, LIBERO-Goal, and LIBERO-Long, where each suite includes 10 language-conditioned manipulation tasks with 50 human demonstrations per task. 
    These suites are designed to probe different types of generalization, including spatial reasoning, object-level transfer, goal-conditioned behavior, and long-horizon compositional manipulation.
    \item \textbf{LIBERO-Plus}~\citep{libero-plus} extends LIBERO with controlled perturbations over seven dimensions, including object layouts, camera viewpoints, robot initial states, language instructions, lighting conditions, background textures, and sensor noise, thereby providing a substantially more challenging testbed for generalization under realistic variations.
    To further examine robustness beyond the standard LIBERO setting, we additionally evaluate the trained model on LIBERO-Plus without re-training. 
    \item \textbf{CALVIN}~\citep{calvin} is a long-horizon language-conditioned manipulation benchmark built on four simulated environments.
    CALVIN contains 34 manipulation tasks and 1,000 language instructions, and evaluates a policy by the average number of successfully completed sub-tasks in multi-step task sequences. 
    Following the standard protocol, we report results on the benchmark's compositional generalization settings to measure long-horizon control and transfer to unseen environments.
    \end{itemize}

    \paragrapht{Backbone and implementation details.}
    We follow the implementation of \citep{openvla} and \citep{spatial-forcing} to build \method on top of OpenVLA-OFT~\citep{openvla-oft}.
    Unlike the original setting, we use Prismatic VLM~\cite{prismatic} with Qwen2.5-0.5B~\citep{qwen25} as the language backbone, and adopt only DINOv2~\citep{dinov2} as the visual encoder without SigLIP~\citep{siglip}.
    On top of this backbone, \method keeps the original visual pathway frozen, augments it with a LoRA-based spatial branch, and introduces a lightweight knowledge router for selecting action-relevant visual tokens.
    We use VGGT~\citep{vggt} as the pre-trained 3D expert, which is kept frozen and used only during training for spatial representation distillation.
    We provide more details in \cref{app:setting}.
\begin{table*}[t]
    \centering
    \footnotesize
    \caption{Comparisons with state-of-the-art methods on LIBERO-Plus. Bold indicates the best performances.
    }\label{tab:libero-plus}
    \setlength{\tabcolsep}{2pt}
        \begin{tabular}{lccccccccccc}
        \toprule
         \multirow{2}[2]{*}{Method}   &  \multirow{2}[2]{*}{Suite}    &   Camera    &   Robot   &   Language    &   Light   &   Background  &   Noise   &   Layout  &   Total   &   \multirow{2}[2]{*}{\makecell{LLM \\ Size}}    &    \multirow{2}[2]{*}{\makecell{Train. \\ Params.}}   \\
          \cmidrule(lr){3-3} \cmidrule(lr){4-4} \cmidrule(lr){5-5} \cmidrule(lr){6-6} \cmidrule(lr){7-7} \cmidrule(lr){8-8} \cmidrule(lr){9-9} \cmidrule(lr){10-10}
             &   &   SR (\%)      &   SR (\%)     &   SR (\%)     &   SR (\%)     &   SR (\%)  &  SR (\%)  & SR (\%)  & SR (\%)  &  &   \\
        \midrule
        \multicolumn{3}{l}{\hspace{-6pt}\gray{\textit{Large-scale models $\geq$ 7B}}} \\ 
        OpenVLA~\citep{openvla}     &   Avg.    &   0.8    &   3.5    &   23.0    &   8.1    &   34.8    &  15.2    &   28.5  &  15.6    &   7B  &  111M  \\
        OpenVLA-OFT~\citep{openvla-oft} &   Avg.    &   56.4    &   31.9    &   79.5    &   88.7    &   93.3    &   75.8    &   74.2    &   69.6    &   7B  & 111M \\
        UniVLA~\citep{univla-iclr}     &   Avg. &   1.8  &   46.2    &   69.6    &   69.0    &   81.0    &   21.2    &   31.9    &   42.9    &   7B  &  8B   \\
        \midrule
        \multicolumn{3}{l}{\hspace{-6pt}\gray{\textit{1B $\leq$ Small-scale models $<$ 7B}}} \\ 
        $\pi_0$~\citep{pi0}     &   Avg.    &   13.8    &   6.0    &   58.8    &   85.0 &   81.4    &   79.0    &   68.9    &   53.6  & 2B &  3.3B \\
        $\pi_0$-FAST~\citep{pi0-fast}    &   Avg.    &   65.1    &   21.6    &   61.0    &   73.2   &   73.2    &   74.4    &   68.8    &   61.6  &  2B  & 3.3B \\
        NORA~\citep{nora}   &   Avg.    &   2.2 &   37.0    &   65.1    &   45.7    &   58.6    &   12.8    &   62.1    &   39.0    & 3B  & 4B  \\
        VLANeXt~\citep{vlanext} &   Avg.   &   \bf 71.6 &   \bf 63.4   &   \bf 81.4   &   93.7   &   91.8   &   89.5   &    77.7    &   \bf 80.1    &   2B  &  2.7B \\ 
        \midrule
        \multicolumn{3}{l}{\hspace{-6pt}\gray{\textit{Tiny-scale models $<$ 1B}}} \\
        VLA-Adapter~\citep{vla-adapter} &   Avg.    &   36.4 &   37.4 &   73.8    &   71.0    &   76.6    &   57.2 &   70.2 &   59.0    &   0.5B    &   598M    \\
        \method (\textbf{Ours}) 
        &   Avg.   &    68.5  &   59.8   & 72.7    &  \bf 94.6   &  \bf 95.9 & \bf 89.6    &  \bf 78.8   &  78.6    &   0.5B    &   53M  \\
        \bottomrule
        \end{tabular}
    \end{table*}

\begin{table*}[t]
    \centering
    \small
    \caption{Comparisons with state-of-the-art methods on CALVIN benchmark. 
    Bold indicates the best performance.
    }\label{tab:calvin}
    \setlength{\tabcolsep}{4pt}
        \begin{tabular}{lccccccccc}
        \toprule
         \multirow{2}[2]{*}{Method} &   \multirow{2}[2]{*}{Task}   &    \multicolumn{5}{c}{Tasks Completed in a Row}    &  \multirow{2}[2]{*}{Avg. Len $\uparrow$}    & \multirow{2}[2]{*}{\makecell{LLM \\ Size}}    &    \multirow{2}[2]{*}{\makecell{Train. \\ Params.}}   \\
         \cmidrule(lr){3-7} 
                &         &   1     &   2     &   3     &   4  &    5   &   &   &  \\
        \midrule
        MCIL~\citep{mcil}   &   ABCD$\rightarrow$D  &   0.377   &   0.027   &   0.002   &   0.000   &   0.000   &   0.40    &   -   &   -   \\
        RT-1~\citep{rt-1}   &   ABCD$\rightarrow$D  &   0.844   &   0.617   &   0.438   &   0.323   &   0.227   &   2.45    &   -   &   -   \\
        Robo-Flamingo~\citep{robo-flamingo} &   ABCD$\rightarrow$D  &   0.964   &   0.896   &   0.824   &   0.740   &   0.660   &   4.09    &   7B   &   1.4B   \\
        GR-1~\citep{gr-1}   &   ABCD$\rightarrow$D  &   0.949   &   0.896   &   0.844   &   0.789   &   0.731   &   4.21    &   -   &   -   \\
        UP-VLA~\citep{up-vla}   &   ABCD$\rightarrow$D  &   0.962   &   0.921   &   0.879   &   0.842   &   0.812   &   4.42    &   -   &   1.5B   \\
        RoboVLMs~\citep{robovlms}   &   ABCD$\rightarrow$D  &   0.967   &   0.930   &   0.899   &   0.865   &   0.826   &   4.49    &   1.3B   &   1.7B   \\
        UniVLA~\citep{univla-iclr}  &   ABCD$\rightarrow$D     &   \bf 0.985    &  \bf 0.961    &   0.931    &   0.899    &   \bf 0.851    &   \bf 4.63    &   7B  &  8B   \\
        \method (\textbf{Ours}) &   ABCD$\rightarrow$D  & 0.979  & 0.960  & \bf 0.934  & \bf 0.900  & 0.843  & 4.61  &   0.5B    &   53M \\
        \midrule
        MCIL~\citep{mcil}   &   ABC$\rightarrow$D  &   0.304   &   0.013   &   0.002   &   0.000   &   0.000   &   0.31    &   -   &   -   \\
        Robo-Flamingo~\citep{robo-flamingo} &   ABC$\rightarrow$D  &   0.824   &   0.619   &   0.466   &   0.331   &   0.235   &   2.47    &   7B   &   1.4B   \\
        SuSIE~\citep{susie} &   ABC$\rightarrow$D  &   0.870   &   0.690   &   0.490   &   0.380   &   0.260   &   2.69    &   -   &   -   \\
        GR-1~\citep{gr-1}   &   ABC$\rightarrow$D  &   0.854   &   0.712   &   0.596   &   0.497   &   0.401   &   3.06    &   -   &   -   \\
        MoDE~\citep{mode}   &   ABC$\rightarrow$D  &   0.962   &   0.889   &   0.811   &   0.718   &   0.635   &   4.01    &   -   &   740M   \\
        UP-VLA~\citep{up-vla}   &   ABC$\rightarrow$D  &   0.928   &   0.865   &   0.815   &   0.769   &   0.699   &   4.08    &   1.3B   &   1.5B   \\ 
        RoboVLMs~\citep{robovlms}   &   ABC$\rightarrow$D  &   0.980   &   0.936   &   0.854   &   0.778   &   0.704   &   4.25    &   1.3B   &   1.7B   \\
        Seer~\citep{seer}   &   ABC$\rightarrow$D  &   0.963   &   0.916   &   0.861   &   0.803   &   0.740   &   4.28    &   -   &   315M   \\
        Dita~\citep{dita}   &   ABC$\rightarrow$D  &   0.945   &   0.825   &   0.728   &   0.613   &   0.500   &   3.61    &   -   &   221M   \\
        UniVLA~\citep{univla-iclr}  &   ABC$\rightarrow$D     &  \bf 0.989    &   0.948    & \bf 0.890    &   0.828    &  \bf 0.751    &   4.41    &   7B  &  8B   \\
        \method (\textbf{Ours}) &   ABC$\rightarrow$D  & 0.985  &  \bf 0.955 & 0.888  & \bf 0.842  & 0.750  & \bf 4.42  &   0.5B    &   53M \\
        \bottomrule
        \end{tabular}
    \end{table*}
\subsection{Results in simulation environment}
    In this section, we compare our \method with state-of-the-art VLAs on simulation benchmarks.
    Please refer to \cref{app:sim} for additional analysis, including the effectiveness of each component of \method, the impact of hyperparameters, and visualization.
    
    \paragrapht{Results on LIBERO.}
    \cref{tab:libero} shows the performance evaluated on the LIBERO benchmark. We categorize the baselines according to the model size: \textit{Large-scale models} with more than 7B parameters, \textit{small-scale models} having parameters between 1B and 7B, and \textit{tiny-scale models} with less than 1B parameters. In addition, we specify the LLM size and the number of trainable parameters.
    \method achieves an average success rate of 98.5\% with only a 0.5B language backbone and 53M trainable parameters, which is the best average performance.
    Compared with VLA-Adapter~\citep{vla-adapter}, \method obtains the same average success rate while requiring about 11$\times$ fewer trainable parameters.
    Comparison with Evo-1~\citep{evo-1} shows \method improves the average success rate by 3.7\% with 14$\times$ fewer trainable parameters, further demonstrating the effectiveness of the proposed method.
    In addition, \method outperforms larger models~\citep{4d-vla, pi0, spatialvla, nora, smolvla, groot, vlanext, octo, openvla, openvla-oft, univla-iclr, cog-vla, spec-prune-vla} despite using a substantially smaller language backbone and far fewer trainable parameters.
    Overall, the results suggest that \method achieves a favorable balance between model scale, trainable parameter efficiency, and task performance in the simulation environment.

    \paragrapht{Results on LIBERO-Plus.}
    Evaluation on LIBERO-Plus measures the robustness of VLAs under diverse visual perturbations.
    As shown in \cref{tab:libero-plus}, 
    \method achieves an average success rate of 78.6\%, substantially outperforming the tiny-scale baseline (\ie, VLA-Adapter~\citep{vla-adapter}) by 19.6\%.
    Moreover, \method is competitive with larger models and achieves the best performance under light, background, noise, and layout perturbations.
    Compared with the performance on LIBERO, the drop is especially small under background and light perturbations, suggesting that the proposed knowledge router effectively suppresses task-irrelevant or noisy visual tokens while preserving action-critical information.
    Although VLANeXt~\citep{vlanext} achieves a slightly higher total performance, it requires a 2B language backbone and 2.7B trainable parameters, while our \method provides comparable robustness with more efficient model configurations.
    We provide per-task and per-perturbation results with additional analysis in \cref{app:libero-plus-app}.

    \paragrapht{Results on CALVIN.}
    In \cref{tab:calvin}, the performance on ABCD$\rightarrow$D shows a general long-horizon language-conditioned manipulation capability, \ie, training the model on the four different environments and evaluating on the environment D. In contrast, the performance on ABC$\rightarrow$D provides the zero-shot evaluation setting in which the models are trained on the environments A, B, and C, and evaluated on the new environment D.
    In ABCD$\rightarrow$D, \method achieves an average length of 4.61, which is comparable to UniVLA~\citep{univla-iclr} with 4.63, while obtaining the best success rates for completing three and four tasks in a row.
    In the more challenging ABC$\rightarrow$D setting, \method achieves the best average length of 4.42 and the best success rates for two- and four-task completion.
    The results show that \method provides strong long-horizon manipulation and zero-shot environment generalization with only a 0.5B language model and 53M trainable parameters.

\section{Real-world Experiments}

\subsection{Experimental setting}

    \begin{figure}
        \centering
        \includegraphics[width=1\linewidth]{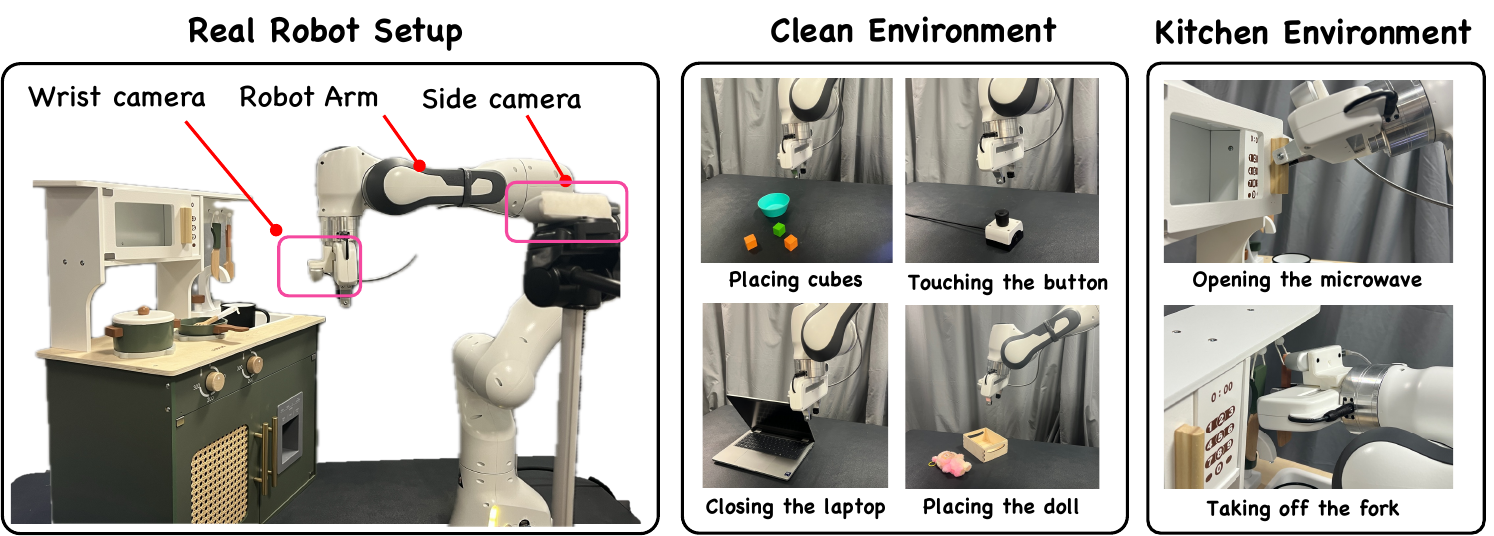}
        \caption{Overview of our real-world robot setup and tasks. We use a 7-DoF robotic arm with two cameras for manipulation. We evaluate six distinct tasks across a clean environment and a more complex kitchen environment.}
        \label{fig:real_robot}
    \end{figure}

\paragrapht{Setup.}
    To evaluate the effectiveness of \method in the real world, we design two scenarios with different levels of complexity.
    The first scenario consists of four manipulation tasks with a clean background: (1) \textit{placing cubes}, (2) \textit{closing the laptop}, (3) \textit{placing the doll}, and (4) \textit{touching the button}.
    These tasks evaluate basic manipulation skills, including pick-and-place and articulated-object interaction.
    We further construct a more challenging kitchen scenario with cluttered objects and task-relevant background distractors.
    As shown in \cref{fig:real_robot}, the robot must localize target objects among distractors and perform spatial reasoning from dual-camera observations.
    This scenario includes two tasks: (1) \textit{opening the microwave} and (2) \textit{taking off the fork}.

   
    We use a 7-DoF Franka Emika Panda robot with a two-finger gripper equipped with a wrist camera and a side camera for visual observation.
    We train the model separately for the two scenarios using a total of 150 human-collected demonstration trajectories recorded at 15Hz.
    Our \method is initialized from the pre-trained Qwen2.5VL~\citep{qwen25}.
    Since this model is not originally trained on robotics-specific data, we follow the training protocol of $\pi_0$~\citep{pi0} and VLANeXt~\citep{vlanext}:
    we first pre-train the model on the DROID dataset~\citep{droid} for 100k steps, and then fine-tune on our real-world demonstrations with LoRAs for 20k steps.
    During evaluation, we measure the performance on each task over 10 episodes and report the average success rate.
    
 \paragraph{Baselines.}
 We compare our \method with two standard VLA baselines:%
\begin{itemize}[leftmargin=*]
    \item 
    
    \textbf{OpenVLA}~\citep{openvla} is a standard 7B VLA model that predicts actions through direct action regression.
    We include OpenVLA as a representative large-scale VLA baseline, as it has shown strong performance on simulated manipulation benchmarks~\citep{libero}.
    
    \item 
    
    \textbf{$\pi_0$-Droid}~\citep{pi0} is a 3B VLA model that combines a vision-language backbone with a flow-matching action expert for continuous control.
    We use the publicly available checkpoint that was finetuned on DROID~\cite{droid}, which contains Franka robot demonstrations and is therefore well-aligned with our real-world setup.
    
\end{itemize}

\subsection{Results in real-world environments}
    
    \cref{fig:real_exp} reports the real-world manipulation performance across six tasks.
    OpenVLA-7B shows limited transfer to our real-world setup, achieving an average success rate of 11.7\%.
    The stronger $\pi_0$-3B baseline improves the average success rate to 45.0\%, especially on relatively clean tasks such as \textit{placing the doll} and \textit{touching a button}.
    In comparison, \method\ achieves the best average success rate of 52.0\% with only a 0.5B language backbone.
    Notably, \method\ improves over $\pi_0$-3B on four out of six tasks, including \textit{closing the laptop}, \textit{touching a button}, \textit{opening the microwave}, and \textit{taking off the knife}.
    The gains on the kitchen tasks are particularly meaningful, as they require target localization under cluttered backgrounds and spatial reasoning from dual-camera observations.
    These results demonstrate that the proposed spatial representation distillation and knowledge routing are effective not only in simulation but also in real-world manipulation.
    


\definecolor{drawVLA}{HTML}{000000}
\definecolor{fillVLA}{HTML}{E6E6E6}

\definecolor{drawPI}{HTML}{1F4E79}
\definecolor{fillPI}{HTML}{CFE2F3}

\definecolor{drawOUR}{HTML}{8B1E1E}
\definecolor{fillOUR}{HTML}{F4CCCC}

\begin{figure}[t]
    \centering
    \begin{tikzpicture}
        \begin{axis}[
            ybar=0pt,
            bar width=10pt,
            width=12cm,
            height=3.5cm,
            ymin=0, ymax=100,
            ytick={0,25,50,75,100},
            ylabel={Success Rate (\%)},
            ylabel style={font=\small},
            ymajorgrids=true,
            grid style={dashed, gray!40},
            axis line style={black, thick},
            tick style={black, thick},
            xtick={1,2,3,4,5,6,7},
            xticklabels={Cubes, Laptop, Doll, Button, Microwave, Fork, Avg.},
            xticklabel style={font=\small},
            tick align=outside,
            enlarge x limits=0.08,
            axis x line*=bottom,
            axis y line*=left,
            legend style={
                at={(0.5,1.12)},
                anchor=south,
                legend columns=-1,
                draw=none,
                font=\small,
                /tikz/every even column/.append style={column sep=0.6cm}
            },
            legend image code/.code={
                \draw[#1] (0pt,-0.6ex) rectangle (0.18cm,0.8ex);
            }
        ]
        
        \addplot[
            fill=fillVLA,
            draw=drawVLA,
            line width=0.6pt,
            pattern=north east lines,
            pattern color=drawVLA
        ] coordinates {
            (1, 0) (2, 20) (3, 0) (4, 50) (5, 0) (6, 0) (7, 11.7)
        };
        
        \addplot[
            fill=fillPI,
            draw=drawPI,
            line width=0.6pt,
            pattern=north east lines,
            pattern color=drawPI
        ] coordinates {
            (1, 30) (2, 30) (3, 90) (4, 80) (5, 20) (6, 20) (7, 45.0)
        };
        
        \addplot[
            fill=fillOUR,
            draw=drawOUR,
            line width=0.6pt,
            pattern=north east lines,
            pattern color=drawOUR
        ] coordinates {
            (1, 20) (2, 50) (3, 80) (4, 90) (5, 40) (6, 30) (7, 52.0)
        };
        
        \legend{OpenVLA-7B, $\pi_{0}$-3B, Ours-0.5B}
        
        \draw[dashed, black!60, line width=0.8pt] (axis cs:6.5, 0) -- (axis cs:6.5, 100);
        
        \end{axis}
    \end{tikzpicture}
    \caption{Performance comparison on real-world tasks.}
    \label{fig:real_exp}
\end{figure}
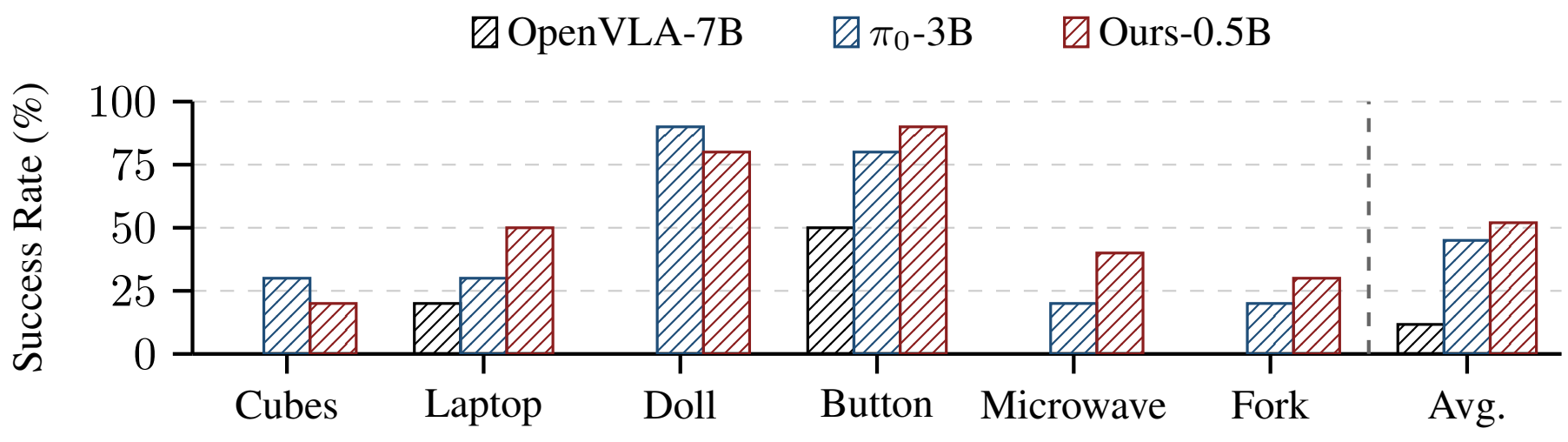
\section{Conclusion}
    We presented \method, a simple and effective framework for enhancing spatial understanding in vision-language-action models. 
    Instead of relying on explicit 3D inputs at inference time or directly reshaping the original visual representation, \method preserves the pre-trained visual pathway and introduces a LoRA-adapted spatial branch learned from a frozen 3D expert through attention response distillation. 
    To effectively integrate generic and spatial cues, we further introduced a lightweight knowledge router that selects instruction- and action-relevant tokens from both branches using gradient-derived supervision. 
    Extensive experiments on simulated and real-world manipulation benchmarks demonstrate that \method achieves strong performance with a compact 0.5B language backbone and a small number of trainable parameters, showing that spatial knowledge can be effectively transferred to VLAs without sacrificing deployment simplicity.


\paragrapht{Broader impact}
    Our work contributes toward more efficient and spatially aware robot policies that can operate without additional 3D sensors during inference. 
    This can reduce hardware dependency and make VLA-based robot learning more accessible across different platforms. 
    By improving spatial reasoning with a compact model, \method may also help lower the computational cost of deploying embodied agents in practical environments. 
    At the same time, more capable robotic systems should be deployed with careful consideration of safety, reliability, and human oversight, especially in environments involving people or fragile objects.

\paragrapht{Limitations}
    Despite its effectiveness, \method still relies on a pre-trained 3D expert during training, and the quality of spatial distillation may depend on the teacher model and the visual diversity of the training data. 
    In addition, the router is trained with pseudo supervision derived from action-loss gradients, which can be noisy when the action model is not sufficiently optimized. 
    We mitigate this issue with warm-up training and momentum-based score accumulation, but more principled routing supervision remains an interesting direction. 
    Finally, although \method shows promising real-world performance, our real-world evaluation is still limited in task diversity and scale. Extending the framework to broader manipulation scenarios, more diverse embodiments, and dynamic human-centered environments is an important future direction.
\section*{Acknowledgment}
This work was supported by the National Research Foundation of Korea(NRF) grant
funded by the Korea government(MSIT) (RS-2025-00515741) and the AI Seoul Tech
Research Support Program of the Seoul Future Foundation.

\newpage
\appendix

\setcounter{figure}{0}
\setcounter{table}{0}

\renewcommand{\thefigure}{\Alph{section}.\arabic{figure}}
\renewcommand{\thetable}{\Alph{section}.\arabic{table}}

\section*{Appendix}\label{app:app}
\section{Simulation Environment}\label{app:sim}

\begin{table*}[h]
    \centering
    \small
    \caption{Hyperparameters for simulated experiments.
    }\label{tab:setting}
    \setlength{\tabcolsep}{4pt}
        \begin{tabular}{ll}
        \toprule
         Hyperparameter   &  Value  \\         
        \midrule
         GPUs    &   NVIDIA RTX A6000 48GB $\times$ 4 \\
         Batch size per GPU &   16  \\
         Initial learning rate / Final learning rate & 5e-5 / 5e-8  \\
         Warm-up with \cref{eq:warm}    &   2k steps    \\
         Learning rate scheduler    &   Cosine annealing lr scheduler   \\ 
         Input images   &   1 exo-view + 1 wrist view   \\
         Image size & 224$\times$224 \\
         LoRA rank  &   32  \\
         Use proprioception &   Yes \\
         Use FiLM   &   No  \\
         Trainable parameters   &   \makecell[l]{53M total: 23.2M (LLM LoRAs) + 12.6M (vision LoRAs) \\ + 4.8M (vision proj.) + 4.3M (knowledge router) \\ + 7.3M (action head) + 0.8M (proprio. proj.) } \\
         Momentum parameter $\beta$ in \cref{eq:momentum}   &   0.9 \\
         $\lambda_\text{attn}$, $\lambda_\text{router}$, $\lambda_\text{bud}$, $\lambda_\text{ent}$ &   1.0, 0.5, 0.05, 0.001\\
        \bottomrule
        \end{tabular}
    \end{table*}

\subsection{Default setting}\label{app:setting}
    \tabref{tab:setting} shows a default setting used in our experiments.

\subsection{Training regime}\label{app:training-detail}
    \paragrapht{Warm-up and joint optimization.}
    Reliable router supervision requires meaningful spatial representations and a reasonably trained action prediction head. 
    We therefore adopt a two-stage training procedure. 
    In the first stage, we warm up the model without the router objective:
    \begin{equation}\label{eq:warm}
        \mathcal{L}_{\text{warm}}
        =
        \mathcal{L}_{\text{act}}
        + \lambda_{\text{attn}} \mathcal{L}_{\text{attn}}.
    \end{equation}
    This stage trains the LoRA-adapted spatial branch with the attention response distillation loss while optimizing the action head with the action loss. 
    After warm-up, we enable the knowledge router and optimize the full objective in \cref{eq:objective}.
    In this stage, the router is trained with pseudo targets derived from action-loss gradients, while the spatial branch continues to receive distillation supervision from the frozen 3D expert. This procedure stabilizes routing training by first establishing geometry-aware visual features and meaningful action gradients, and then learning to select tokens that are useful for action generation.

    \paragrapht{Inference without 3D experts.}
    At inference time, the pre-trained 3D expert is completely removed. 
    Given an RGB observation $\mathbf{I}_t$ and a language instruction, \method extracts the generic visual representation $\mathbf{Z}_v$ from the original visual encoder and the spatial representation $\mathbf{Z}_p$ from the LoRA-adapted branch. 
    The knowledge router predicts token-wise gate scores conditioned on the visual and language features, selects the top-ranked tokens according to the target keeping ratio, and fuses the selected generic and spatial tokens before forwarding them to the language model for action prediction. 
    Thus, \method benefits from spatial knowledge distilled during training while requiring neither depth maps, point clouds, nor external 3D experts at inference time.


\subsection{Component analysis}\label{app:component}
    In this section, we further analyze the impact of each component in our \method, including LoRA layers in the LLM and the vision encoder, projector to connect the visual tokens to the LLM, the usage of the vision encoder, attention distillation, and knowledge routing. In \cref{tab:component}, we compare the performance and the number of trainable parameters according to each component. Note that the action head with $\ell_1$ regression is identically trained on all cases unless we specify.
    
    \paragrapht{Baseline.} We start from the simplest LoRA tuning by applying the LoRA layers only to the LLM in Prismatic Qwen2.5-0.5B~\citep{qwen25}. The results in \cref{tab:component}-(a) show that applying the LoRAs only to the LLM is insufficient to adapt the pre-trained VLMs to manipulation, showing less than 5\% of the average success rate.

    \paragrapht{LoRAs to the vision encoder.} We apply the additional LoRA layers to the vision encoder, \ie, DINOv2~\citep{dinov2} and SigLIP~\citep{siglip}, managing about 28.7M additional trainable parameters. Similar to the baseline, \cref{tab:component}-(b) suggests applying the LoRAs to the model fails to adapt the pre-trained VLM to the manipulation tasks.

    \paragrapht{Projector.} In addition to the LoRAs, training the projector (with 4.8M parameters) to project the visual token embeddings to the LLM's input space yields meaningful results, showing 72.4\% of the average success rate. Note that we train a new projector from scratch instead of re-using the projector in the pre-trained VLM.

    \paragrapht{Usage of vision encoder.} In the original prismatic VLM, two vision encoders are employed to complement with each other for two different types of visual information. Specifically, DINOv2~\citep{dinov2} and SigLIP~\citep{siglip} provide localization-oriented and language-aligned visual information, respectively~\citep{prismatic}. Therefore, removing SigLIP from the pre-trained VLM weakens its language-aligned semantic grounding. In practice, the model without SigLIP performs worse than using both vision encoders, suggesting two complementary information is essential for instruction following manipulation, as shown in \cref{tab:component}-(d).

    \paragrapht{Attention distillation.} We transfer the spatial representations using attention response distillation from VGGT~\citep{vggt} to the LoRA-adapted path in the DINOv2. \cref{tab:component}-(e) shows that injecting spatial knowledge into the pre-trained vision encoder significantly improves the performance of (d) and even outperforms (c), which employs two vision encoders, with only a single vision encoder. This result suggests the consistent conclusion with the recent 3D VLAs~\citep{evo-1, spatial-forcing}.

    \paragrapht{Knowledge routing.} While attention response distillation enables the model to produce the generic visual and spatial representations, redundant or noisy information may be included. Our full model with the knowledge router trained using the action loss-driven supervision feeds forward the instruction- and action-relevant visual tokens to the LLM, yielding the best performance, as shown in \cref{tab:component}-(f). 

    Consequently, all components in our \method are carefully designed to complement each other, requiring minimum additional trainable parameters.
    
    \paragrapht{Objectives.} We now replace $\ell_1$ regression with alternatives for action generation, \ie, the OpenVLA-style next-token prediction and the $\pi_0$-style diffusion policy.
    Compared to regression, next-token prediction reformulates action generation as discrete token modeling, making it naturally compatible with autoregressive language models.
    With next token prediction, the performance can be affected by action discretization errors and the limited resolution of the action vocabulary.

    \begin{wrapfigure}{r}{0.4\textwidth}
    \centering
    \small
    \vspace{-10pt}
    \includegraphics[width=0.4\textwidth]{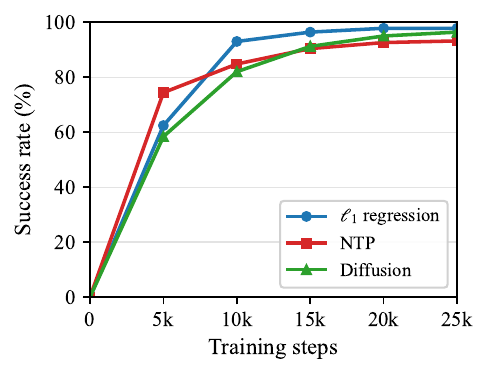}
    \caption{Performance according to the action generation objective. NTP and Diffusion denote next token prediction and diffusion policy, respectively.}
    \label{fig:objective}
    \vspace{-15pt}
    \end{wrapfigure}
    
    Diffusion policy, on the other hand, directly models continuous action trajectories and can capture multi-modal distributions through an iterative denoising process, which is beneficial for ambiguous or long-horizon manipulation, but introduces additional sampling cost at inference time.
    
    In \cref{fig:objective}, we depict the performance evaluated on LIBERO-Goal according to the training steps. 
    We evaluate the model every 5k training steps and report the success rate (\%).
    Next-token prediction shows stronger early learning, achieving 74.4\% at 5k iterations, thanks to its autoregressive nature, while the performance saturates at 93.2\%.
    Diffusion policy starts lower at 58.4\%, but improves steadily as training proceeds and reaches 96.4\% at 25k iterations, benefiting from continuous trajectory modeling and its ability to capture multi-modal action distributions.
    Overall, \(\ell_1\) regression converges fastest and achieves the best final performance in our setting, while next-token prediction favors early optimization and diffusion provides stronger late-stage continuous control.

\begin{table*}[t]
    \centering
    \small
    \caption{Ablation study for each component in our \method.
    }\label{tab:component}
    \setlength{\tabcolsep}{4pt}
        \begin{tabular}{lcccccc}
        \toprule
         \multirow{2}[2]{*}{Method}   &  Spatial    &   Object  &   Goal    &   Long    &   Average     &    \multirow{2}[2]{*}{\makecell{Train. \\ Params.}}   \\
         \cmidrule(lr){2-2} \cmidrule(lr){3-3} \cmidrule(lr){4-4} \cmidrule(lr){5-5} \cmidrule(lr){6-6}
                &   SR (\%)      &   SR (\%)     &   SR (\%)     &   SR (\%)     &   SR (\%)  &  \\
        \midrule
        (a) only LLM LoRAs  & -  & -  & - & - & <\;5\% &   31.4M   \\
        (b) +\;Vision LoRAs  & -  &  - & - & - & <\;5\% &   60.1M   \\
        (c) +\;Projector & 82.6 & 84.4 & 77.4 & 45.0 & 72.4 &   64.9M    \\
        (d) -\;SigLIP &  62.4  &   76.8   &   58.4   &   37.2  &  58.7  &   48.8M    \\
        (e) +\;Attention Distill. &  95.8  &   96.0   &  94.0    &  90.6   & 94.1  &    48.8M    \\
        (f) +\;Knowledge Router &  99.6  &   99.6   &    98.0  &  96.6   & 98.5  &   53.1M    \\
        \bottomrule
        \end{tabular}
    \end{table*}

\paragrapht{Spatial supervision.}
    We investigate different strategies for transferring geometric knowledge from the pre-trained 3D expert to the spatial pathway. 
    Specifically, we consider four alternatives to our attention-response distillation: (1) action-only adaptation; (2) feature matching; (3) relational matching; and (4) attention-map matching. 
    Action-only adaptation learns spatial representations solely from the downstream action objective, without explicit geometric supervision. 
    Feature matching directly aligns the student and teacher embeddings, while relational matching transfers pairwise similarities between visual tokens. 
    Attention-map matching instead aligns the query-key attention distributions, capturing where spatial information is aggregated. 
    Unlike these approaches, our attention response distillation additionally transfers the teacher's aggregated value representations, providing geometry-aware contextual information conditioned on the VLA queries.

    \begin{table}[t]
        \centering
        \caption{Comparison with different spatial supervision strategies on LIBERO-Goal.}
        \label{tab:distillation}
        \begin{tabular}{lc}
        \toprule
        Spatial supervision & Success rate (\%)\\
        \midrule
        Action loss only & 81.6\\
        Feature matching & 94.8\\
        Relational matching & 95.6\\
        Attention-map matching & 96.4\\
        \method \textbf{(ours)} & \textbf{98.0}\\
        \bottomrule
        \end{tabular}
    \end{table}

    We compare these objectives on LIBERO-Goal while keeping the remaining architecture and training configurations unchanged. 
    As shown in~\cref{tab:distillation}, action-only adaptation achieves a success rate of 81.6\%, whereas feature and relational matching improve the performance to 94.8\% and 95.6\%, respectively. 
    Attention-map matching further increases the success rate to 96.4\%, indicating the benefit of transferring query-conditioned relationships beyond static feature alignment. 
    Our attention-response distillation achieves the highest success rate of 98.0\%, outperforming feature matching by 3.2\% and attention-map matching by 1.6\%. 
    These results suggest that transferring both the attention distribution and its aggregated spatial content is more effective than supervising either latent representations or attention patterns alone.

    \paragrapht{Knowledge fusion.}
    We explore different strategies for integrating generic visual representations with spatially specialized features. 
    Specifically, we compare our action-driven knowledge router with two alternative fusion mechanisms: direct concatenation and feature-wise linear modulation (FiLM)~\citep{film}. 
    Direct concatenation retains all tokens from both pathways and combines their representations without distinguishing their relevance to the downstream task. 
    FiLM-based fusion instead uses the generic features to predict channel-wise affine transformations of the spatial features, allowing the two representations to interact through feature modulation. 
    However, neither approach explicitly determines which knowledge source is more useful at each spatial position. 
    In contrast, our knowledge router selects tokens from the generic and spatial pathways according to their estimated contribution to action prediction, enabling task-dependent integration of complementary visual information.

    As shown in \cref{tab:fusion}, direct concatenation achieves a success rate of 94.0\%, while FiLM-based fusion slightly improves the performance to 94.8\%. 
    Our action-driven router further increases the success rate to 98.0\%, outperforming the two alternatives by 4.0 and 3.2 percentage points, respectively. 
    Interestingly, combining all available features does not necessarily lead to better action prediction, suggesting that generic and spatial representations may contain redundant or task-irrelevant information. 
    These results highlight the importance of selectively integrating the two knowledge sources, rather than uniformly combining their representations.

    \begin{table}[t]
        \centering
        \caption{Comparison with different knowledge fusion strategies on LIBERO-Goal.}
        \label{tab:fusion}
        \begin{tabular}{lc}
        \toprule
        Fusion strategy & Success rate (\%)\\
        \midrule
        Direct all-token concatenation & 94.0\\
        FiLM-based branch fusion & 94.8\\
        \textbf{Action-driven knowledge router (ours)} & \textbf{98.0}\\
        \bottomrule
        \end{tabular}
    \end{table}

\subsection{Hyperparameters}\label{app:hyperparameter}
    In this section, we provide additional analysis corresponding to the hyperparameters in our \method.
    Note that all experiments in this section are conducted on the LIBERO-Goal suite.

    \paragrapht{Momentum-based moving average.}
    In our \method, only routed tokens are forwarded to the LLM, making tokens that are not selected in the current training step cannot directly affect the action loss and thus do not provide reliable instantaneous gradient signals.
    In other words, the supervision would be biased toward the currently routed subset, as the routing targets were derived only from the current gradients. 
    To mitigate this issue, we maintain a momentum-based moving average of the token importance scores:
    \begin{equation}\label{eq:momentum}
        \bar{s}_n \leftarrow \beta \bar{s}_n + (1-\beta)s_n,
    \end{equation}
    where $\beta$ is the momentum coefficient and $s_n=\mathrm{ReLU}(-\partial \mathcal{L}_{\mathrm{act}}/\partial p_n)$ is the instantaneous importance score. 
    The accumulated score $\bar{s}_n$ provides a stable estimate of token utility across training steps, allowing previously informative tokens to remain considered even when they are not selected in the current step. 
    We use $\bar{\mathbf{s}}=\{\bar{s}_n\}_{n=1}^{N}$ to construct the routing target by top-$k$ selection.

    \begin{table*}[h]
    \centering
    \small
    \caption{Performance according to the momentum parameter $\beta$ on LIBERO-goal.
    }\label{tab:momentum}
    \setlength{\tabcolsep}{4pt}
        \begin{tabular}{lccccc}
        \toprule
         Momentum   &  0.0    &   0.9 &   0.99    &   0.999   &   0.9999 \\
        \midrule
        Avg. SR (\%)    &   94.0   &  98.0    &   97.6   &   97.2   &   97.0    \\
        \bottomrule
        \end{tabular}
    \end{table*}

    In \cref{tab:momentum}, we compare the performance evaluated on the LIBERO-Goal suite according to the value of the momentum parameter $\beta$.
    With $\beta=0.0$, the router is trained using only the instantaneous importance scores, resulting in a lower success rate of 94.0\%.
    This indicates that routing supervision becomes unstable when it depends solely on the gradients from the currently selected tokens.
    Introducing momentum substantially improves performance, achieving the best success rate of 98.0\% with $\beta=0.9$.
    This demonstrates that accumulating the token importance scores across training steps provides more reliable routing targets.
    However, overly large momentum values slightly degrade performance, as the accumulated scores are updated too slowly and may not sufficiently reflect the current policy and scene distribution.
    We therefore use $\beta=0.9$ as the default setting.

\begin{figure}[t]
        \centering
        \begin{subfigure}[b]{0.325\linewidth}
            \includegraphics[width=\linewidth]{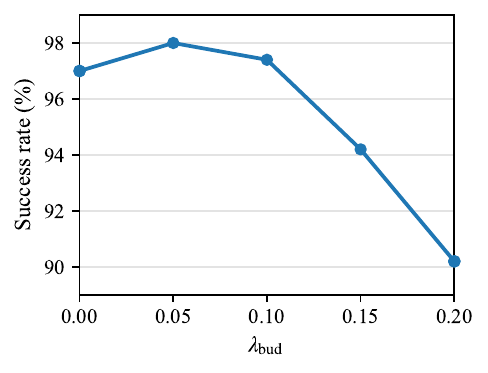}
            \caption{Effectiveness of $\lambda_\text{budget}$}
            \label{fig:lambda_bud}
        \end{subfigure}
        \hfill
        \begin{subfigure}[b]{0.325\linewidth}
            \includegraphics[width=\linewidth]{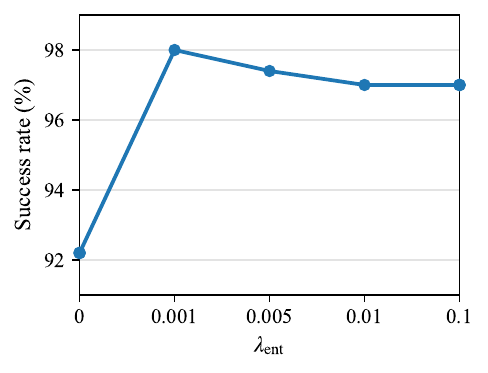}
            \vspace{-1.0em}
            \caption{Effectiveness of $\lambda_\text{ent}$}
            \label{fig:lambda_ent}
        \end{subfigure}
        \hfill
        \begin{subfigure}[b]{0.325\linewidth}
            \includegraphics[width=\linewidth]{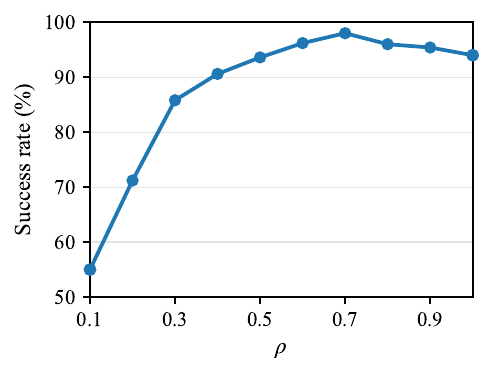}
            \vspace{-1.0em}
            \caption{Effectiveness of $\rho$}
            \label{fig:rho}
        \end{subfigure}
        \caption{
        Ablation study for each hyperparameter. We report the performance evaluated on LIBERO-Object according to the values of (a) the budget coefficient $\lambda_\text{bud}$, (b) the entropy coefficient $\lambda_\text{ent}$, and the target keeping ratio $\rho$. }
        \label{fig:hyperparameter}
    \end{figure}

    \paragrapht{Coefficient $\lambda_\text{bud}$.} 
    The budget loss encourages the average router gate probability to match the target routing ratio $\rho$.
    In \cref{fig:lambda_bud}, we evaluate different values of $\lambda_\text{bud}$ from 0.0 to 0.2 with an interval of 0.05 on the LIBERO-Goal suite.
    As $\lambda_\text{bud}$ increases, the optimization places more emphasis on satisfying the global routing budget, which reduces the flexibility of token selection.
    As a result, the router is encouraged to match the average gate probability to $\rho$ rather than to preserve locally important tokens for each instruction and action.
    This overly strict budget constraint suppresses useful deviations that could otherwise be controlled by entropy regularization, leading to degraded performance.
    In particular, increasing $\lambda_\text{bud}$ to 0.2 reduces the success rate to 90.2\%.

    \paragrapht{Coefficient $\lambda_\text{ent}$.} 
    The entropy loss regularizes the router to prevent gate probability collapse and maintain adaptive token selection.
    We report the performance with different values of $\lambda_\text{ent}$ in \cref{fig:lambda_ent}.
    When $\lambda_\text{ent}=0$, the router is regularized only by the budget loss, resulting in a lower success rate of 92.2\%, as discussed above.
    Introducing entropy regularization stabilizes routing by preventing early saturation of gate probabilities and improving robustness to noisy pseudo targets from gradient-based supervision.
    As $\lambda_\text{ent}$ increases, the entropy term encourages more confident, near-binary gate decisions.
    This sharpening is beneficial at moderate levels, but an overly large $\lambda_\text{ent}$ makes the routing decisions too rigid and reduces the flexibility of token selection during training.
    We obtain the best performance with $\lambda_\text{ent}=0.001$.
    

    \paragrapht{Target keeping ratio $\rho$.} 
    The target keeping ratio $\rho$ controls the proportion of visual tokens forwarded to the LLM.
    The model completely fails when $\rho=0.0$, since no visual information is provided for action prediction.
    As shown in \cref{fig:rho}, increasing $\rho$ consistently improves performance from 55.0\% at $\rho=0.1$ to 96.2\% at $\rho=0.6$, indicating that sufficient visual context is necessary for reliable manipulation.
    The best performance is achieved at $\rho=0.7$, reaching 98.0\%, which suggests that preserving a moderately large but selective set of tokens provides the best balance between visual sufficiency and noise suppression.
    When $\rho$ is further increased, the success rate gradually decreases to 96.0\%, 95.4\%, and 94.0\% at $\rho=0.8$, $0.9$, and $1.0$, respectively.
    This drop shows that forwarding excessive tokens weakens the effect of routing, as redundant or instruction-irrelevant representations can interfere with action prediction.

\subsection{Additional analysis on LIBERO-Plus}\label{app:libero-plus-app}
    \begin{table*}[t]
    \centering
    \footnotesize
    \caption{Per-task and per-perturbation performance evaluated on LIBERO-Plus. Bold indicates the best performances.
    }\label{tab:libero-plus-app}
    \setlength{\tabcolsep}{2pt}
        \begin{tabular}{lccccccccccc}
        \toprule
         \multirow{2}[2]{*}{Method}   &  \multirow{2}[2]{*}{Suite}    &   Camera    &   Robot   &   Language    &   Light   &   Background  &   Noise   &   Layout  &   Total   &   \multirow{2}[2]{*}{\makecell{LLM \\ Size}}    &    \multirow{2}[2]{*}{\makecell{Train. \\ Params.}}   \\
          \cmidrule(lr){3-3} \cmidrule(lr){4-4} \cmidrule(lr){5-5} \cmidrule(lr){6-6} \cmidrule(lr){7-7} \cmidrule(lr){8-8} \cmidrule(lr){9-9} \cmidrule(lr){10-10}
             &   &   SR (\%)      &   SR (\%)     &   SR (\%)     &   SR (\%)     &   SR (\%)  &  SR (\%)  & SR (\%)  & SR (\%)  &  &   \\
        \midrule
        \multicolumn{3}{l}{\hspace{-6pt}\gray{\textit{Large-scale models $\geq$ 7B}}} \\ 
        \multirow{5}{*}{OpenVLA-OFT~\citep{openvla-oft}} &   Spatial    &   88.3    &   40.0     & 80.5 & 98.3    &   97.3    &   96.3    &   93.9    &   84.0    &   \multirow{5}{*}{7B}  & \multirow{5}{*}{111M} \\
        &   Object  &   38.9    &   25.4     & 99.0 & 73.7    &   97.6    &   72.3    &   71.8    &   66.5    & &   \\
        &   Goal    &   62.0    &   25.2     & 53.2 & \bf 93.9    &   \bf 92.5    &   75.2    &   59.1    &   63.0    & &   \\
        &   Long    &   38.7    &   38.2     & 87.0 & 89.4    &   86.8    &   63.5    &   76.9    &   66.4    & &   \\ \cmidrule{2-10}
        &   Avg.    &   56.4    &   31.9     & 79.5 & 88.7    &   93.3    &   75.8    &   74.2    &   69.6    & &   \\
        \midrule
        \multicolumn{3}{l}{\hspace{-6pt}\gray{\textit{1B $\leq$ Small-scale models $<$ 7B}}} \\ 
        \multirow{5}{*}{VLANeXt~\citep{vlanext}} &   Spatial    &   \bf 94.4    &   \bf 83.4     & \bf 86.7 & \bf 99.3    &   98.4    &   98.0    &   92.2    &   \bf 92.8  &     \multirow{5}{*}{2B}  &  \multirow{5}{*}{2.7B} \\
        &   Object  &   \bf 73.2    &   \bf 67.8     & \bf 84.2 & 98.3    &   94.8    &   92.2    &   73.0    &   \bf 82.1    & &   \\
        &   Goal    &   \bf 71.1    &   \bf 50.6     & \bf 66.3 & 92.5    &   81.9    &   \bf 90.2    &   66.8    &   \bf 72.7    & &   \\
        &   Long    &   47.7    &   \bf 51.9     & 88.5 & 84.7    &   92.0    &   77.5    &   78.8    &   72.8    & &   \\ \cmidrule{2-10}
        &   Avg.   &   \bf 71.6 &   \bf 63.4   &   \bf 81.4   &   93.7   &   91.8   &   89.5   &    77.7    &   \bf 80.1    &   &   \\ 
        \midrule
        \multicolumn{3}{l}{\hspace{-6pt}\gray{\textit{Tiny-scale models $<$ 1B}}} \\
        \multirow{5}{*}{\method (\textbf{Ours})} &   Spatial   &    92.6  &   82.0   & 76.9    &  \bf 99.3   &  \bf 99.2 & \bf 98.3    &  \bf 93.2   &  90.1 &   \multirow{5}{*}{0.5B}    &   \multirow{5}{*}{53B}    \\   %
        &   Object   &    69.4  &   63.6   & 78.5    &  \bf 99.3   &  \bf 98.4 & \bf 94.3    &  \bf 75.2   &  81.3     \\
        &   Goal   &    66.2  &   45.7   & 58.5    &  84.2   &  85.4 & \bf 90.2    &  \bf 68.2   &  70.3     \\
        &   Long   &    \bf 48.2  &   50.9   & 78.1    &  \bf 94.9   &  \bf 94.8 & \bf 78.0    &  \bf 80.1   &  \bf 72.9     \\ \cmidrule{2-10}
        &   Avg.   &    68.5  &   59.8   & 72.7    &  \bf 94.6   &  \bf 95.9 & \bf 89.6    &  \bf 78.8   &  78.6     \\
        \bottomrule
        \end{tabular}
    \end{table*}

    \cref{tab:libero-plus-app} provides per-task and per-perturbation results on LIBERO-Plus.
    Overall, \method achieves strong robustness across all four suites with only a 0.5B language backbone and 53M trainable parameters.
    On the Spatial suite, \method obtains 90.1\% total success rate and achieves the best performance under light, background, noise, and layout perturbations.
    This suggests that the proposed spatial representation distillation and knowledge routing are particularly effective when manipulation requires robust spatial reasoning under visual changes.

    \begin{figure}[t]
        \centering
        \includegraphics[width=1\linewidth]{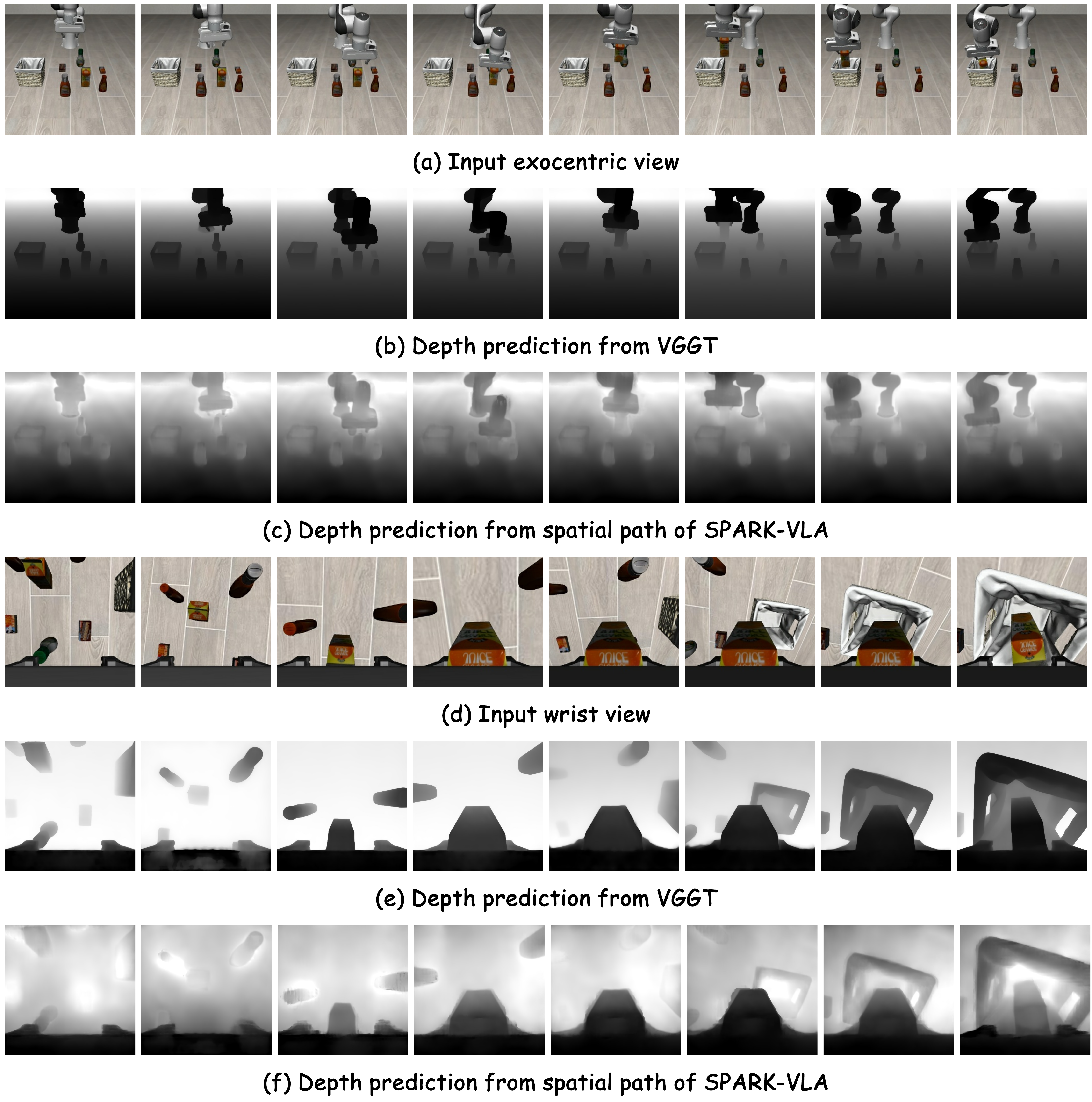}
        \caption{Qualitative analysis of learned spatial representations. From given exocentric and wrist view observations, we extract depth maps using the original VGGT and the spatial path in our \method.}
        \label{fig:depth}
    \end{figure}
    
    On the Object and Goal suites, \method remains competitive with larger models, although VLANeXt or OpenVLA-OFT shows stronger performance under some camera, robot, and language perturbations.
    In contrast, \method consistently performs well under appearance-level perturbations such as light, background, and noise.
    For example, on the Object suite, \method achieves the best success rates under light, background, noise, and layout perturbations, despite using substantially fewer trainable parameters than VLANeXt.
    On the Long suite, \method achieves the best total success rate of 72.9\%, slightly outperforming VLANeXt, and obtains the best performance under camera, light, background, noise, and layout perturbations.
    These results indicate that selectively routing task-relevant generic and spatial tokens is especially beneficial for long-horizon manipulation, where irrelevant visual changes can accumulate errors across multiple action steps.
    
    In summary, while larger models can be more robust to certain perturbation types such as language and robot appearance changes, \method shows strong robustness to visual perturbations that affect scene appearance and spatial layout.
    This supports our main observation that knowledge routing helps suppress task-irrelevant visual variations while preserving action-critical evidence.

    \begin{figure}[t]
        \centering
        \includegraphics[width=1\linewidth]{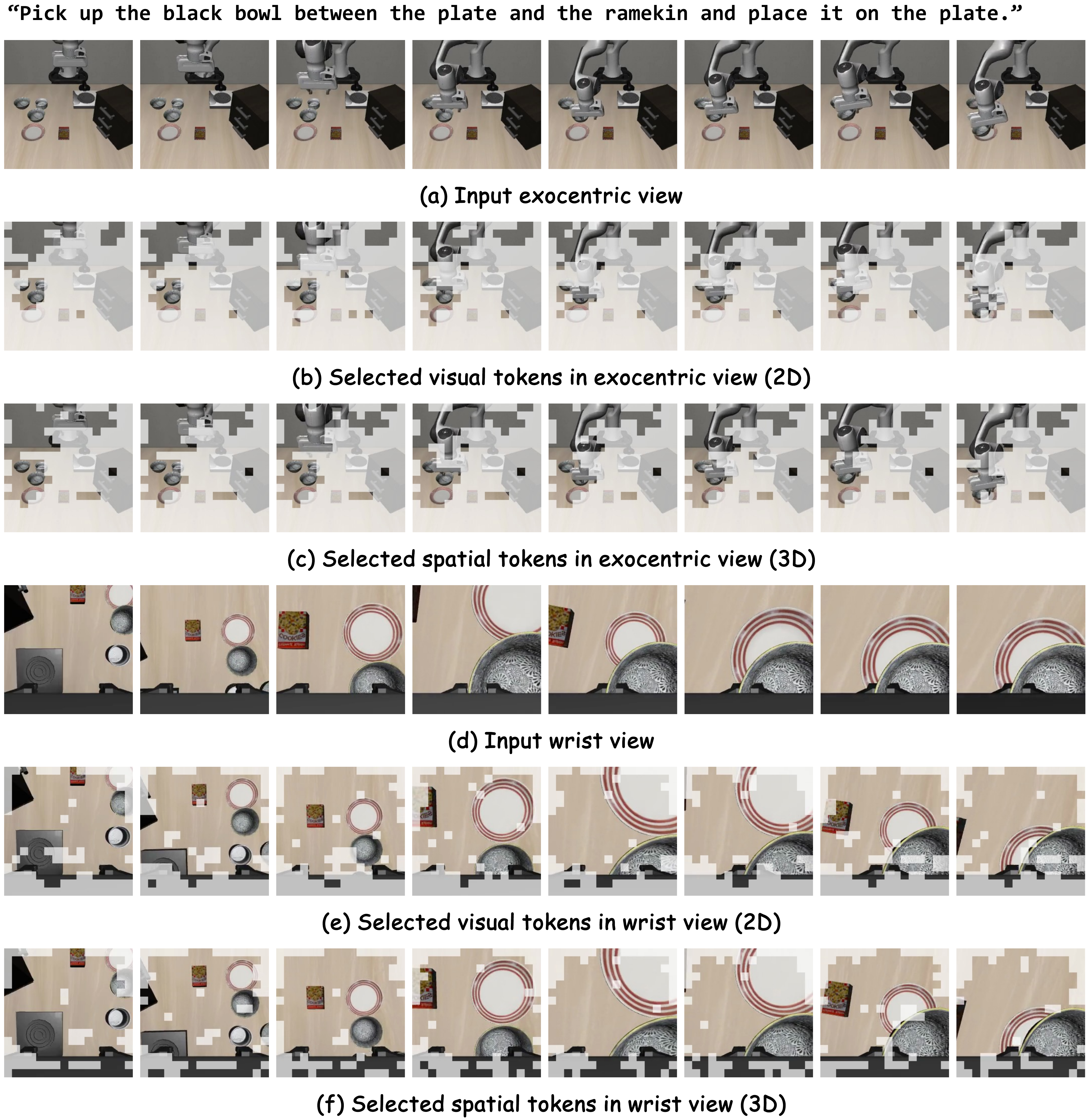}
        \caption{Qualitative analysis of selected tokens. From given exocentric and wrist view observations, the knowledge router in our \method selectively forwards instruction- and action-relevant visual (2D) and spatial (3D) tokens to the LLM. We mask out the tokens, which are not selected in the router.}
        \label{fig:token}
    \end{figure}

\subsection{Visualization}\label{app:visualization_sim}
    \paragrapht{Does spatial representation distillation indeed transfer 3D information into the vision encoder?}
    We present a new implicit adaptation strategy to inject spatial knowledge to the pre-trained vision encoder.
    To examine this question, we use VGGT~\citep{vggt} as a diagnostic model:
    we replace its original DINOv2 encoder with our LoRA-adapted vision encoder, while keeping the remaining components of VGGT fixed, and predict depth maps from the resulting representations.
    If the adapted encoder has acquired spatial knowledge, the modified VGGT should still recover meaningful scene geometry despite the encoder replacement. 
    As shown in \cref{fig:depth}, the predicted depth maps preserve the global scene structure and object boundaries, demonstrating that the adapted encoder captures geometry-aware representations through the proposed attention response distillation. 
    This suggests that spatial representation distillation injects useful 3D understanding into the vision encoder, rather than merely improving action prediction through task-specific adaptation.

    \paragrapht{Which tokens are preserved and excluded in the knowledge router?}
    We further analyze the behavior of the proposed knowledge router to examine which visual tokens are preserved for action generation.
    Specifically, we visualize (a) the original exocentric video and the selected tokens from the exocentric view in (b) the original visual and (c) spatial branches.
    We also visualize (d) the original wrist view video and the selected tokens from the wrist view in (e) the original visual and (f) spatial branches.
    As shown in \cref{fig:token}, the router exhibits view-dependent selection behavior.
    In the exocentric view, it preserves only a sparse subset of tokens around task-relevant objects, interaction regions, and spatial context, while discarding large background areas.
    This is because the exocentric camera provides a wide and relatively stable scene-level view, where many tokens correspond to static background or redundant global context.

    In contrast, the router selects a denser set of tokens from the wrist view.
    Unlike the exocentric view, the wrist camera is closely coupled with the end-effector and directly observes fine-grained manipulation cues, such as object pose, gripper-object contact, local geometry, and occlusion changes.
    Therefore, a larger portion of wrist view tokens can contribute to action prediction, especially when precise control depends on local spatial evidence near the manipulated object.
    This indicates that the router does not simply prune tokens uniformly across views, but adaptively allocates the visual budget according to action utility.
    Across both views, the original branch tends to preserve semantically informative regions, while the spatial branch emphasizes geometrically relevant areas around objects and contact regions.
    These visualizations suggest that knowledge routing selectively balances sparse contextual evidence from the exocentric view with dense local manipulation evidence from the wrist view.

\section{Real-World Environment}

    \begin{figure}[!t]
        \centering
        \includegraphics[width=1\linewidth]{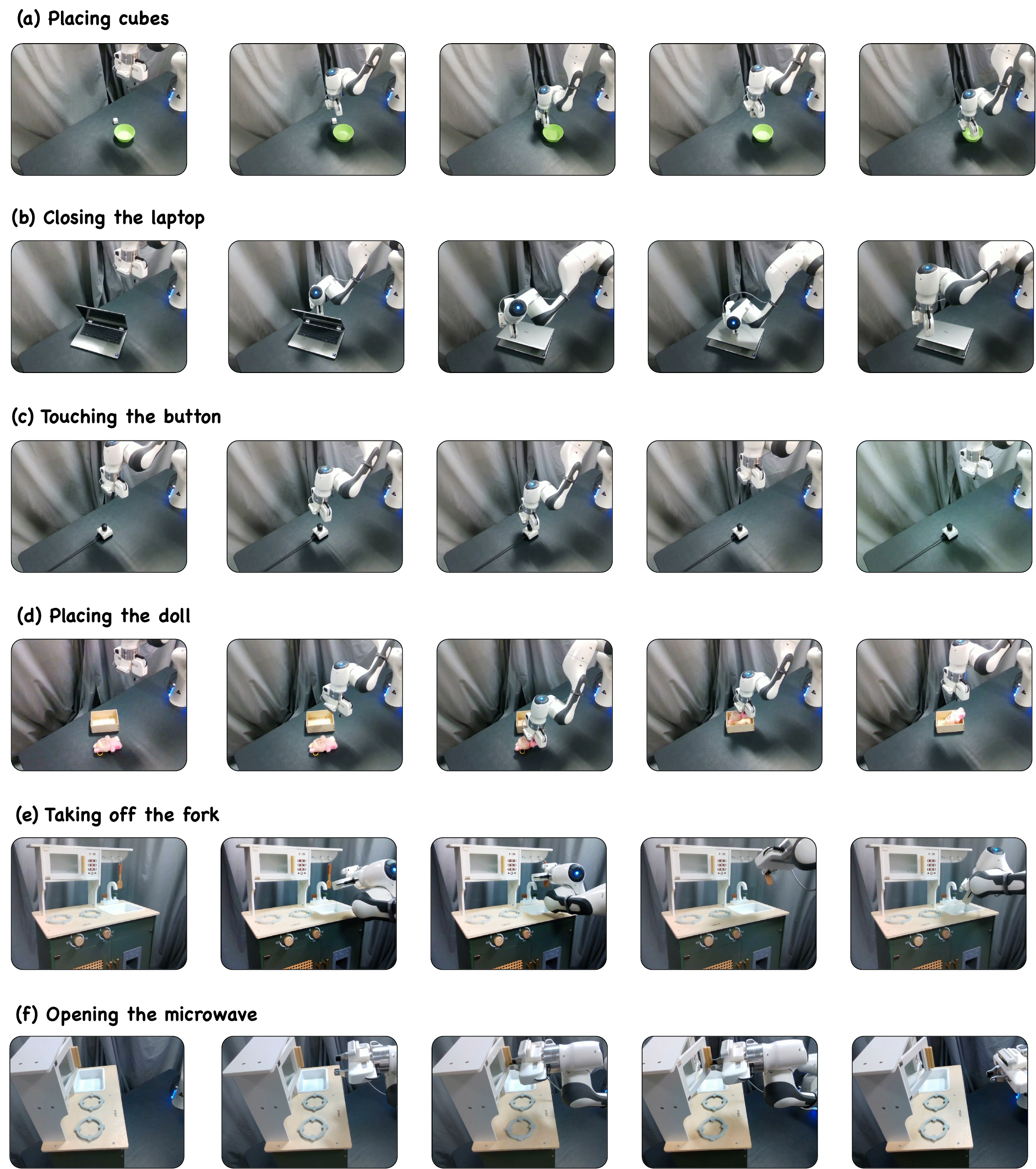}
        \caption{Qualitative experiments of our method in real-world tasks.}
        \label{fig:real_robot_visulization}
    \end{figure}

We present demos of our real-world experiments for the six tasks mentioned in the main paper, showing the execution process in Figure~\ref{fig:real_robot_visulization}. We also provide addtional real-world experiments and analysis in the following sections.

\subsection{Additional real-world experiments}
\label{sec:additional_real_world}
To further assess the effectiveness of SPARK-VLA across a broader range of manipulation tasks, we extend the real-robot evaluation with four additional kitchen manipulation tasks: (1) \textit{stacking bowls}, (2) \textit{pouring water into the sink}, (3) \textit{placing a pan in the sink}, and  (4) \textit{picking up a rag}. These tasks involve multi-object spatial alignment, container manipulation, precise placement, and deformable-object interaction, expanding our real-world evaluation from six tasks to ten. For these additional experiments, we use a different hareware of Robotiq two-finger gripper and fine-tune both methods for 30k steps, following a training protocol similar to that described in the main paper. Each task is evaluated over 10 trials per method.

\begin{figure}[htbp]
    \centering
    \begin{minipage}[t]{0.64\linewidth}
        \vspace{0pt}
        \centering
        \definecolor{drawPI}{HTML}{1F4E79}
\definecolor{fillPI}{HTML}{CFE2F3}
\definecolor{drawOUR}{HTML}{8B1E1E}
\definecolor{fillOUR}{HTML}{F4CCCC}

\begin{tikzpicture}
    \begin{axis}[
        ybar=0pt,
        bar width=12pt,
        width=\linewidth,
        height=4.8cm,
        ymin=0, ymax=100,
        ytick={0,25,50,75,100},
        ylabel={Success Rate (\%)},
        ylabel style={font=\small},
        ymajorgrids=true,
        grid style={dashed, gray!40},
        axis line style={black, thick},
        tick style={black, thick},
        xtick={1,2,3,4,5},
        xticklabels={
            \shortstack{Stacking \\ bowls},
            \shortstack{Placing \\ pan},
            \shortstack{Pouring \\ water},
            \shortstack{Picking \\ up rag},
            Avg.
        },
        xticklabel style={font=\small, align=center},
        tick align=outside,
        enlarge x limits=0.12,
        axis x line*=bottom,
        axis y line*=left,
        legend style={
            at={(0.5,1.12)},
            anchor=south,
            legend columns=-1,
            draw=none,
            font=\small,
            /tikz/every even column/.append style={column sep=0.6cm}
        },
        legend image code/.code={
            \draw[#1] (0pt,-0.6ex) rectangle (0.18cm,0.8ex);
        }
    ]
        \addplot[
            fill=fillPI,
            draw=drawPI,
            line width=0.6pt,
            postaction={pattern=north east lines, pattern color=drawPI}
        ] coordinates {
            (1,30) (2,10) (3,80) (4,100) (5,55.0)
        };

        \addplot[
            fill=fillOUR,
            draw=drawOUR,
            line width=0.6pt,
            postaction={pattern=north east lines, pattern color=drawOUR}
        ] coordinates {
            (1,50) (2,30) (3,80) (4,100) (5,65.0)
        };

        \legend{$\pi_0$-3B, SPARK-VLA}
        \draw[dashed, black!60, line width=0.8pt]
            (axis cs:4.5,0) -- (axis cs:4.5,100);
    \end{axis}
\end{tikzpicture}
    \end{minipage}\hfill
    \begin{minipage}[t]{0.32\linewidth}
        \vspace{0pt}
        \centering
        \includegraphics[width=\linewidth]{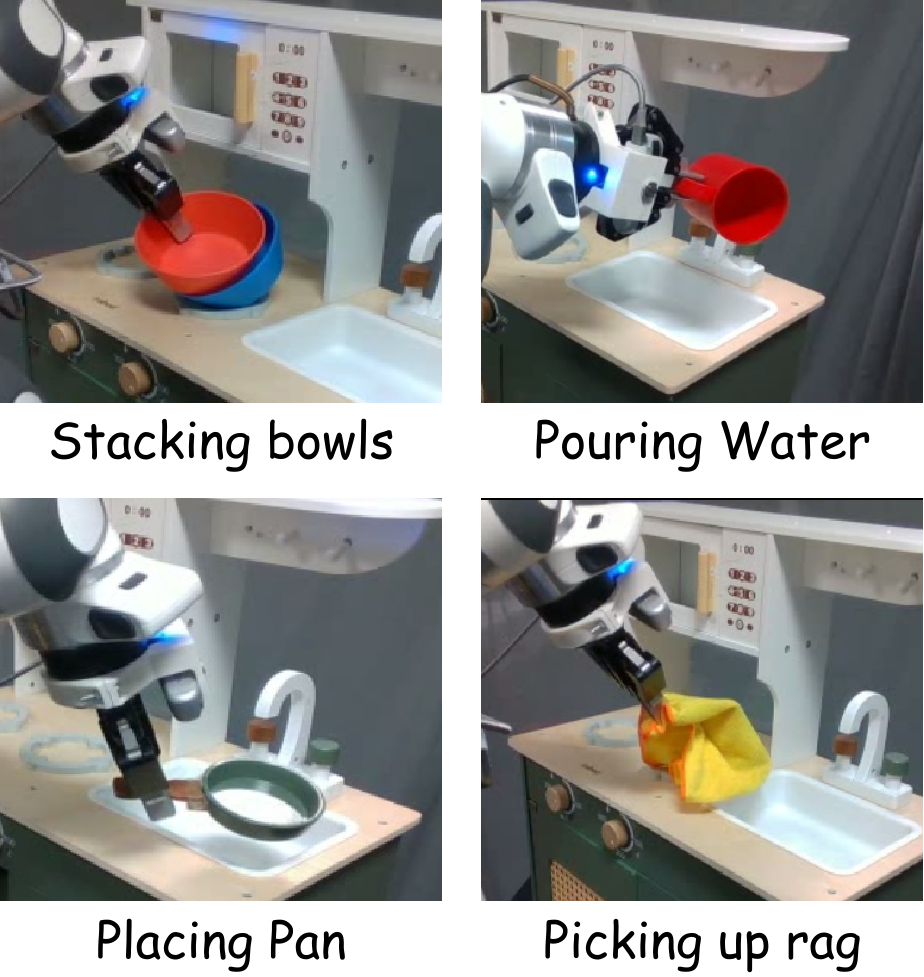}
    \end{minipage}
    \caption{Success rates on four additional kitchen manipulation tasks (left) and additional real-world examples (right). The success-rate plot reports 10 trials per task per method; Avg. denotes the mean across the four tasks.}
    \label{fig:additional_real_exp}
\end{figure}

As shown in the Figure~\ref{fig:additional_real_exp}, SPARK-VLA achieves an average success rate of 65.0\%, compared with 55.0\% for \(\pi_0\)-3B, with gains of 20\% on both stacking bowls and placing the pan in the sink, which require more precise reasoning about relative object position, orientation, and target geometry.

\subsection{Generalization to unseen objects and language instructions}
\label{sec:generalization}

To test whether VGGT-based spatial distillation contributes to semantic generalization, we conducted additional real-world experiments to evaluate SPARK-VLA under previously unseen objects and language instructions. We compare SPARK-VLA with the same architecture trained without VGGT-based spatial distillation, denoted as SPARK w/o VGGT. We evaluate the same pick-and-place behavior-grasping a target object and placing it in the sing using one object (seen during training) and five unseen objectrs. The unseen objects vary in color, shape, handle structure, and semantic category. For reproducibility, each object was tested at the same five pre-defined starting locations. Across all six objects, SPARK-VLA succeeds in 18 of 30 trials (60.0\%), compared with 14 of 30 (46.7\%) for SPARK w/o VGGT. 

\begin{table}[t]
    \centering
    \caption{Pick-and-place success rates for SPARK-VLA and SPARK w/o VGGT on the training red cup and five held-out objects. Each object is tested at the same five predefined starting locations (five trials per object).}
    \label{tab:unseen_objects}
    \small
    \begin{tabular}{@{}p{0.38\columnwidth}p{0.3\columnwidth}cc@{}}
        \toprule
        Object & Condition & SPARK-VLA & w/o VGGT \\
        \midrule
        Red cup & Training object & 80\% & 60\% \\
        Blue cup & Unseen color & 40\% & 40\% \\
        Black cup & Unseen color & 60\% & 40\% \\
        Green handleless paper cup & Unseen color and geometry & 40\% & 40\% \\
        Small milk carton & Unseen category and geometry & 60\% & 60\% \\
        Hand-cream bottle & Unseen category and geometry & 80\% & 40\% \\
        \midrule
        \textbf{Mean} & Across six objects & \textbf{60.0\%} & {46.7\%} \\
        \bottomrule
    \end{tabular}
\end{table}

\begin{table}[!t]
    \centering
    \caption{Outcomes for SPARK-VLA and SPARK w/o VGGT on nine language instructions held out from task training. Complete success requires executing the requested manipulation with the correct target; partial success denotes a correct grasp without completing the requested placement.}
    \label{tab:unseen_instructions}
    \small
    \begin{tabular}{@{}p{0.70\columnwidth}cc@{}}
        \toprule
        Language instruction & SPARK-VLA & w/o VGGT \\
        \midrule
        ``Pick the cup in the sink.'' & Complete & Complete \\
        ``Move the cup to the sink.'' & Complete & Failure \\
        ``Pick the cup and place it in the right-side area.'' & Complete & Failure \\
        ``Move the red object to the right-side area.'' & Partial & Complete \\
        ``Get the cup ready to wash in the sink.'' & Partial & Failure \\
        ``Pick up the handleless green cup and put it in the sink.'' & Complete & Complete \\
        ``Move the blue food can to the sink.'' & Failure & Failure \\
        ``Pick up the white hand-cream bottle.'' & Complete & Complete \\
        ``Pick up the blue-and-orange Capri-Sun juice pouch and put it in the sink.'' & Complete & Failure \\
        \midrule
        \textbf{Complete success} & \textbf{6/9 (66.7\%)} & 4/9 (44.4\%) \\
        \bottomrule
    \end{tabular}
\end{table}

We further evaluated nine language instructions that were not used during training. They include unseen paraphrases, spatial attributes, object descriptions, and category names. SPARK-VLA achieves six complete successes and two partial successes, whereas SPARK w/o VGGT achieves four complete successes and no partial successes. These results show that SPARK-VLA can execute several previously unseen object descriptions and linguistic formulations. However, they should not be interpreted as evidence that geometric distillation increases the semantic capacity of the language model itself.

Our failure analysis instead suggests that VGGT-based distillation primarily improves geometric execution after the target object has been identified. In many failures of SPARK w/o VGGT, the robot approached the correct object but closed the gripper too early or too late. We therefore hypothesize that the distilled spatial representation improves object-proximity estimation, grasp timing, and placement accuracy across changes in object appearance and shape. In this sense, the geometric prior complements the pretrained model's semantic knowledge rather than replacing it.



\bibliography{main}

@String(CVPR= {IEEE Conf. Comput. Vis. Pattern Recog.})

@String(ICCV= {Int. Conf. Comput. Vis.})

@String(ECCV= {Eur. Conf. Comput. Vis.})

@String(NIPS= {Adv. Neural Inform. Process. Syst.})

@String(ICLR = {Int. Conf. Learn. Represent.})

@String(AAAI = {AAAI})

@String(CVPR  = {CVPR})

@String(ICCV  = {ICCV})

@String(ECCV  = {ECCV})

@String(NIPS  = {NeurIPS})

@String(ICML  = {ICML})

@String(ICLR  = {ICLR})

@String(TMLR  = {TMLR})

@String(CoRL  = {CoRL})

@String(RSS  = {RSS})

@String(ICRA  = {ICRA})

@article{evo-0,
  title={Evo-0: Vision-language-action model with implicit spatial understanding},
  author={Lin, Tao and Li, Gen and Zhong, Yilei and Zou, Yanwen and Du, Yuxin and Liu, Jiting and Gu, Encheng and Zhao, Bo},
  journal={arXiv preprint arXiv:2507.00416},
  year={2025}
}

@inproceedings{evo-1,
  title={Evo-1: Lightweight vision-language-action model with preserved semantic alignment},
  author={Lin, Tao and Zhong, Yilei and Du, Yuxin and Zhang, Jingjing and Liu, Jiting and Chen, Yinxinyu and Gu, Encheng and Liu, Ziyan and Cai, Hongyi and Zou, Yanwen and others},
  booktitle=CVPR,
  year={2026}
}

@article{openvla,
  title={Openvla: An open-source vision-language-action model},
  author={Kim, Moo Jin and Pertsch, Karl and Karamcheti, Siddharth and Xiao, Ted and Balakrishna, Ashwin and Nair, Suraj and Rafailov, Rafael and Foster, Ethan and Lam, Grace and Sanketi, Pannag and others},
  journal={arXiv preprint arXiv:2406.09246},
  year={2024}
}

@article{spatial-forcing,
  title={Spatial forcing: Implicit spatial representation alignment for vision-language-action model},
  author={Li, Fuhao and Song, Wenxuan and Zhao, Han and Wang, Jingbo and Ding, Pengxiang and Wang, Donglin and Zeng, Long and Li, Haoang},
  journal={arXiv preprint arXiv:2510.12276},
  year={2025}
}

@inproceedings{vggt,
  title={Vggt: Visual geometry grounded transformer},
  author={Wang, Jianyuan and Chen, Minghao and Karaev, Nikita and Vedaldi, Andrea and Rupprecht, Christian and Novotny, David},
  booktitle=CVPR,
  year={2025}
}

@inproceedings{lora,
  title={Lora: Low-rank adaptation of large language models.},
  author={Hu, Edward J and Shen, Yelong and Wallis, Phillip and Allen-Zhu, Zeyuan and Li, Yuanzhi and Wang, Shean and Wang, Lu and Chen, Weizhu and others},
  booktitle=ICLR,
  year={2022}
}

@article{qwen25,
  title={Qwen2.5-vl technical report},
  author={Bai, Shuai and Chen, Keqin and Liu, Xuejing and Wang, Jialin and Ge, Wenbin and Song, Sibo and Dang, Kai and Wang, Peng and Wang, Shijie and Tang, Jun and others},
  journal={arXiv preprint arXiv:2502.13923},
  year={2025}
}

@inproceedings{rt-2,
  title={Rt-2: Vision-language-action models transfer web knowledge to robotic control},
  author={Zitkovich, Brianna and Yu, Tianhe and Xu, Sichun and Xu, Peng and Xiao, Ted and Xia, Fei and Wu, Jialin and Wohlhart, Paul and Welker, Stefan and Wahid, Ayzaan and others},
  booktitle=CoRL,
  year={2023},
}

@article{pi0,
  title={$\pi_0$: A Vision-Language-Action Flow Model for General Robot Control},
  author={Black, Kevin and Brown, Noah and Driess, Danny and Esmail, Adnan and Equi, Michael and Finn, Chelsea and Fusai, Niccolo and Groom, Lachy and Hausman, Karol and Ichter, Brian and others},
  journal={arXiv preprint arXiv:2410.24164},
  year={2024}
}

@article{dexvla,
  title={Dexvla: Vision-language model with plug-in diffusion expert for general robot control},
  author={Wen, Junjie and Zhu, Yichen and Li, Jinming and Tang, Zhibin and Shen, Chaomin and Feng, Feifei},
  journal={arXiv preprint arXiv:2502.05855},
  year={2025}
}

@inproceedings{cot-vla,
  title={Cot-vla: Visual chain-of-thought reasoning for vision-language-action models},
  author={Zhao, Qingqing and Lu, Yao and Kim, Moo Jin and Fu, Zipeng and Zhang, Zhuoyang and Wu, Yecheng and Li, Zhaoshuo and Ma, Qianli and Han, Song and Finn, Chelsea and others},
  booktitle=CVPR,
  year={2025}
}

@article{spatialvla,
  title={Spatialvla: Exploring spatial representations for visual-language-action model},
  author={Qu, Delin and Song, Haoming and Chen, Qizhi and Yao, Yuanqi and Ye, Xinyi and Ding, Yan and Wang, Zhigang and Gu, JiaYuan and Zhao, Bin and Wang, Dong and others},
  journal={arXiv preprint arXiv:2501.15830},
  year={2025}
}

@article{tracevla,
  title={Tracevla: Visual trace prompting enhances spatial-temporal awareness for generalist robotic policies},
  author={Zheng, Ruijie and Liang, Yongyuan and Huang, Shuaiyi and Gao, Jianfeng and Daum{\'e} III, Hal and Kolobov, Andrey and Huang, Furong and Yang, Jianwei},
  journal={arXiv preprint arXiv:2412.10345},
  year={2024}
}

@article{3d-cavla,
  title={3d cavla: Leveraging depth and 3d context to generalize vision language action models for unseen tasks},
  author={Bhat, Vineet and Lan, Yu-Hsiang and Krishnamurthy, Prashanth and Karri, Ramesh and Khorrami, Farshad},
  journal={arXiv preprint arXiv:2505.05800},
  year={2025}
}

@article{geovla,
  title={Geovla: Empowering 3d representations in vision-language-action models},
  author={Sun, Lin and Xie, Bin and Liu, Yingfei and Shi, Hao and Wang, Tiancai and Cao, Jiale},
  journal={arXiv preprint arXiv:2508.09071},
  year={2025}
}

@inproceedings{chatvla-2,
  title={Chatvla-2: Vision-language-action model with open-world reasoning},
  author={Zhou, Zhongyi and Zhu, Yichen and Liu, Xiaoyu and Tang, Zhibin and Wen, Junjie and Peng, Yaxin and Shen, Chaomin and Xu, Yi},
  booktitle=NIPS,
  year=2025
}

@inproceedings{octo,
  title={Octo: An open-source generalist robot policy},
  author={Ghosh, Dibya and Walke, Homer and Pertsch, Karl and Black, Kevin and Mees, Oier and Dasari, Sudeep and Hejna, Joey and Kreiman, Tobias and Xu, Charles and others},
  booktitle=RSS,
  year={2024}
}

@inproceedings{dita,
  title={Dita: Scaling diffusion transformer for generalist vision-language-action policy},
  author={Hou, Zhi and Zhang, Tianyi and Xiong, Yuwen and Duan, Haonan and Pu, Hengjun and Tong, Ronglei and Zhao, Chengyang and Zhu, Xizhou and Qiao, Yu and Dai, Jifeng and others},
  booktitle=ICCV,
  year={2025}
}

@inproceedings{diffusion-policy,
	title={Diffusion Policy: Visuomotor Policy Learning via Action Diffusion},
	author={Chi, Cheng and Feng, Siyuan and Du, Yilun and Xu, Zhenjia and Cousineau, Eric and Burchfiel, Benjamin and Song, Shuran},
	booktitle=RSS,
	year={2023}
}

@inproceedings{pi0-fast,
  title={Fast: Efficient action tokenization for vision-language-action models},
  author={Pertsch, Karl and Stachowicz, Kyle and Ichter, Brian and Driess, Danny and Nair, Suraj and Vuong, Quan and Mees, Oier and Finn, Chelsea and Levine, Sergey},
  booktitle=RSS,
  year={2025}
}

@inproceedings{openvla-oft,
  title={Fine-tuning vision-language-action models: Optimizing speed and success},
  author={Kim, Moo Jin and Finn, Chelsea and Liang, Percy},
  booktitle=RSS,
  year={2025}
}

@inproceedings{vla-adapter,
  title={Vla-adapter: An effective paradigm for tiny-scale vision-language-action model},
  author={Wang, Yihao and Ding, Pengxiang and Li, Lingxiao and Cui, Can and Ge, Zirui and Tong, Xinyang and Song, Wenxuan and Zhao, Han and Zhao, Wei and Hou, Pengxu and others},
  booktitle=AAAI,
  year={2026}
}

@inproceedings{4d-vla,
  title={4d-vla: Spatiotemporal vision-language-action pretraining with cross-scene calibration},
  author={Zhang, Jiahui and Chen, Yurui and Xu, Yueming and Huang, Ze and Zhou, Yanpeng and Yuan, Yu-Jie and Cai, Xinyue and Huang, Guowei and Quan, Xingyue and Xu, Hang and others},
  booktitle=NIPS,
  year={2025}
}

@article{smolvla,
  title={Smolvla: A vision-language-action model for affordable and efficient robotics},
  author={Shukor, Mustafa and Aubakirova, Dana and Capuano, Francesco and Kooijmans, Pepijn and Palma, Steven and Zouitine, Adil and Aractingi, Michel and Pascal, Caroline and Russi, Martino and Marafioti, Andres and others},
  journal={arXiv preprint arXiv:2506.01844},
  year={2025}
}

@article{groot,
  title={Gr00t n1: An open foundation model for generalist humanoid robots},
  author={Bjorck, Johan and Casta{\~n}eda, Fernando and Cherniadev, Nikita and Da, Xingye and Ding, Runyu and Fan, Linxi and Fang, Yu and Fox, Dieter and Hu, Fengyuan and Huang, Spencer and others},
  journal={arXiv preprint arXiv:2503.14734},
  year={2025}
}

@article{nora,
  title={Nora: A small open-sourced generalist vision language action model for embodied tasks},
  author={Hung, Chia-Yu and Sun, Qi and Hong, Pengfei and Zadeh, Amir and Li, Chuan and Tan, U and Majumder, Navonil and Poria, Soujanya and others},
  journal={arXiv preprint arXiv:2504.19854},
  year={2025}
}

@inproceedings{seer,
  title={Predictive inverse dynamics models are scalable learners for robotic manipulation},
  author={Tian, Yang and Yang, Sizhe and Zeng, Jia and Wang, Ping and Lin, Dahua and Dong, Hao and Pang, Jiangmiao},
  booktitle=ICLR,
  year={2025}
}

@inproceedings{univla-iclr,
  title={Unified vision-language-action model},
  author={Wang, Yuqi and Li, Xinghang and Wang, Wenxuan and Zhang, Junbo and Li, Yingyan and Chen, Yuntao and Wang, Xinlong and Zhang, Zhaoxiang},
  booktitle=ICLR,
  year={2026}
}

@inproceedings{palme,
  title={Palm-e: An embodied multimodal language model},
  author={Driess, Danny and Xia, Fei and Sajjadi, Mehdi SM and Lynch, Corey and Chowdhery, Aakanksha and Ichter, Brian and Wahid, Ayzaan and Tompson, Jonathan and Vuong, Quan and Yu, Tianhe and others},
  booktitle=ICML,
  year={2023}
}

@inproceedings{openx,
  title={Open x-embodiment: Robotic learning datasets and rt-x models: Open x-embodiment collaboration 0},
  author={O’Neill, Abby and Rehman, Abdul and Maddukuri, Abhiram and Gupta, Abhishek and Padalkar, Abhishek and Lee, Abraham and Pooley, Acorn and Gupta, Agrim and Mandlekar, Ajay and Jain, Ajinkya and others},
  booktitle=ICRA,
  year={2024},
}

@article{geoaware-vla,
  title={Geoaware-vla: Implicit geometry aware vision-language-action model},
  author={Abouzeid, Ali and Mansour, Malak and Sun, Qinbo and Sun, Zezhou and Song, Dezhen},
  journal={arXiv preprint arXiv:2509.14117},
  year={2025}
}

@inproceedings{vla-cache,
  title={VLA-Cache: Efficient Vision-Language-Action Manipulation via Adaptive Token Caching},
  author={Xu, Siyu and Wang, Yunke and Xia, Chenghao and Zhu, Dihao and Huang, Tao and Xu, Chang},
  booktitle=NIPS,
  year={2025}
}

@article{sp-vla,
  title={Sp-vla: A joint model scheduling and token pruning approach for vla model acceleration},
  author={Li, Ye and Meng, Yuan and Sun, Zewen and Ji, Kangye and Tang, Chen and Fan, Jiajun and Ma, Xinzhu and Xia, Shutao and Wang, Zhi and Zhu, Wenwu},
  journal={arXiv preprint arXiv:2506.12723},
  year={2025}
}

@inproceedings{cog-vla,
  title={Cogvla: Cognition-aligned vision-language-action model via instruction-driven routing \& sparsification},
  author={Li, Wei and Zhang, Renshan and Shao, Rui and He, Jie and Nie, Liqiang},
  booktitle=NIPS,
  year={2025}
}

@inproceedings{semantic-vla,
  title={Semanticvla: Semantic-aligned sparsification and enhancement for efficient robotic manipulation},
  author={Li, Wei and Zhang, Renshan and Shao, Rui and Fang, Zhijian and Zhou, Kaiwen and Tian, Zhuotao and Nie, Liqiang},
  booktitle=AAAI,
  year={2026}
}

@article{spec-prune-vla,
  title={Specprune-vla: Accelerating vision-language-action models via action-aware self-speculative pruning},
  author={Wang, Hanzhen and Xu, Jiaming and Xiang, Yushun and Pan, Jiayi and Zhou, Yongkang and Li, Yong-Lu and Dai, Guohao},
  journal={arXiv preprint arXiv:2509.05614},
  year={2025}
}

@article{vla-pruner,
  title={VLA-Pruner: Temporal-Aware Dual-Level Visual Token Pruning for Efficient Vision-Language-Action Inference},
  author={Liu, Ziyan and Chen, Yeqiu and Cai, Hongyi and Lin, Tao and Yang, Shuo and Liu, Zheng and Zhao, Bo},
  journal={arXiv preprint arXiv:2511.16449},
  year={2025}
}

@inproceedings{libero,
  title={Libero: Benchmarking knowledge transfer for lifelong robot learning},
  author={Liu, Bo and Zhu, Yifeng and Gao, Chongkai and Feng, Yihao and Liu, Qiang and Zhu, Yuke and Stone, Peter},
  booktitle=NIPS,
  year={2023}
}

@article{libero-plus,
  title={Libero-plus: In-depth robustness analysis of vision-language-action models},
  author={Fei, Senyu and Wang, Siyin and Shi, Junhao and Dai, Zihao and Cai, Jikun and Qian, Pengfang and Ji, Li and He, Xinzhe and Zhang, Shiduo and Fei, Zhaoye and others},
  journal={arXiv preprint arXiv:2510.13626},
  year={2025}
}

@article{calvin,
  title={Calvin: A benchmark for language-conditioned policy learning for long-horizon robot manipulation tasks},
  author={Mees, Oier and Hermann, Lukas and Rosete-Beas, Erick and Burgard, Wolfram},
  journal={IEEE Robotics and Automation Letters},
  volume={7},
  number={3},
  pages={7327--7334},
  year={2022},
  publisher={IEEE}
}

@inproceedings{prismatic,
  title={Prismatic vlms: Investigating the design space of visually-conditioned language models},
  author={Karamcheti, Siddharth and Nair, Suraj and Balakrishna, Ashwin and Liang, Percy and Kollar, Thomas and Sadigh, Dorsa},
  booktitle=ICML,
  year={2024}
}

@article{dinov2,
  title={Dinov2: Learning robust visual features without supervision},
  author={Oquab, Maxime and Darcet, Timoth{\'e}e and Moutakanni, Th{\'e}o and Vo, Huy and Szafraniec, Marc and Khalidov, Vasil and Fernandez, Pierre and Haziza, Daniel and Massa, Francisco and El-Nouby, Alaaeldin and others},
  journal=TMLR,
  year={2024}
}

@inproceedings{siglip,
  title={Sigmoid loss for language image pre-training},
  author={Zhai, Xiaohua and Mustafa, Basil and Kolesnikov, Alexander and Beyer, Lucas},
  booktitle=ICCV,
  year={2023}
}

@article{mcil,
  title={Language conditioned imitation learning over unstructured data},
  author={Lynch, Corey and Sermanet, Pierre},
  journal={arXiv preprint arXiv:2005.07648},
  year={2020}
}

@article{rt-1,
  title={Rt-1: Robotics transformer for real-world control at scale},
  author={Brohan, Anthony and Brown, Noah and Carbajal, Justice and Chebotar, Yevgen and Dabis, Joseph and Finn, Chelsea and Gopalakrishnan, Keerthana and Hausman, Karol and Herzog, Alex and Hsu, Jasmine and others},
  journal={arXiv preprint arXiv:2212.06817},
  year={2022}
}

@inproceedings{robo-flamingo,
  title={Vision-language foundation models as effective robot imitators},
  author={Li, Xinghang and Liu, Minghuan and Zhang, Hanbo and Yu, Cunjun and Xu, Jie and Wu, Hongtao and Cheang, Chilam and Jing, Ya and Zhang, Weinan and Liu, Huaping and others},
  booktitle=ICLR,
  year={2024}
}

@inproceedings{gr-1,
  title={Unleashing large-scale video generative pre-training for visual robot manipulation},
  author={Wu, Hongtao and Jing, Ya and Cheang, Chilam and Chen, Guangzeng and Xu, Jiafeng and Li, Xinghang and Liu, Minghuan and Li, Hang and Kong, Tao},
  booktitle=ICLR,
  year={2024}
}

@inproceedings{up-vla,
  title={Up-vla: A unified understanding and prediction model for embodied agent},
  author={Zhang, Jianke and Guo, Yanjiang and Hu, Yucheng and Chen, Xiaoyu and Zhu, Xiang and Chen, Jianyu},
  booktitle=ICML,
  year={2025}
}

@article{robovlms,
  title={Towards generalist robot policies: What matters in building vision-language-action models},
  author={Liu, Huaping and Li, Xinghang and Li, Peiyan and Liu, Minghuan and Wang, Dong and Liu, Jirong and Kang, Bingyi and Ma, Xiao and Kong, Tao and Zhang, Hanbo},
  journal={arXiv preprint arXiv:2412.14058},
  year={2025}
}

@article{susie,
  title={Zero-shot robotic manipulation with pretrained image-editing diffusion models},
  author={Black, Kevin and Nakamoto, Mitsuhiko and Atreya, Pranav and Walke, Homer and Finn, Chelsea and Kumar, Aviral and Levine, Sergey},
  journal={arXiv preprint arXiv:2310.10639},
  year={2023}
}

@article{mode,
  title={Efficient diffusion transformer policies with mixture of expert denoisers for multitask learning},
  author={Reuss, Moritz and Pari, Jyothish and Agrawal, Pulkit and Lioutikov, Rudolf},
  journal={arXiv preprint arXiv:2412.12953},
  year={2024}
}

@inproceedings{focal,
  title={Focal loss for dense object detection},
  author={Lin, Tsung-Yi and Goyal, Priya and Girshick, Ross and He, Kaiming and Doll{\'a}r, Piotr},
  booktitle=ICCV,
  year={2017}
}

@article{vlanext,
  title={VLANeXt: Recipes for Building Strong VLA Models},
  author={Wu, Xiao-Ming and Fan, Bin and Liao, Kang and Jiang, Jian-Jian and Yang, Runze and Luo, Yihang and Wu, Zhonghua and Zheng, Wei-Shi and Loy, Chen Change},
  journal={arXiv preprint arXiv:2602.18532},
  year={2026}
}

@article{droid,
  title={Droid: A large-scale in-the-wild robot manipulation dataset},
  author={Khazatsky, Alexander and Pertsch, Karl and Nair, Suraj and Balakrishna, Ashwin and Dasari, Sudeep and Karamcheti, Siddharth and Nasiriany, Soroush and Srirama, Mohan Kumar and Chen, Lawrence Yunliang and Ellis, Kirsty and others},
  journal={arXiv preprint arXiv:2403.12945},
  year={2024}
}

@inproceedings{v-lynx,
  title={V-LynX: Token interface alignment for video+x llms},
  author={Park, Jungin and Lee, Jiyoung and Sohn, Kwanghoon},
  booktitle=ICML,
  year={2026}
}

@inproceedings{film,
  title     = {{FiLM}: Visual Reasoning with a General Conditioning Layer},
  author    = {Perez, Ethan and Strub, Florian and de Vries, Harm and Dumoulin, Vincent and Courville, Aaron},
  booktitle = {AAAI},
  year      = {2018},
}

@inproceedings{mvsnet,
  title     = {{MVSNet}: Depth Inference for Unstructured Multi-view Stereo},
  author    = {Yao, Yao and Luo, Zixin and Li, Shiwei and Fang, Tian and Quan, Long},
  booktitle = ECCV,
  year      = {2018},
}

@inproceedings{epipolar-transformers,
  title     = {Epipolar Transformers},
  author    = {He, Yihui and Yan, Rui and Fragkiadaki, Katerina and Yu, Shoou-I},
  booktitle = CVPR,
  year      = {2020}
}

@inproceedings{rvt,
  title     = {{RVT}: Robotic View Transformer for 3D Object Manipulation},
  author    = {Goyal, Ankit and Xu, Jie and Guo, Yijie and Blukis, Valts and Chao, Yu-Wei and Fox, Dieter},
  booktitle = {CoRL},
  year      = {2023},
}
\bibliographystyle{plain}

\end{document}